\documentclass[11pt, a4paper, logo, internal]{dm}
\usepackage{graphicx}%
\usepackage{multirow}%
\usepackage{amsmath,amssymb,amsfonts}%
\usepackage{amsthm}%
\usepackage{mathrsfs}%
\usepackage[title]{appendix}%
\usepackage{xcolor}%
\usepackage{textcomp}%
\usepackage{manyfoot}%
\usepackage{booktabs}%
\usepackage{algorithm}%
\usepackage{algorithmicx}%
\usepackage{algpseudocode}%
\usepackage{listings}%
\usepackage{hyperref}
\usepackage{lineno}

\theoremstyle{thmstyleone}%
\theoremstyle{thmstyletwo}%

\theoremstyle{thmstylethree}%
\usepackage{mathtools}
\usepackage[dvipsnames]{xcolor}
\usepackage[numbers, sort&compress, round]{natbib}
\usepackage{booktabs}
\usepackage{graphicx}
\usepackage{xfrac}
\usepackage{bbm}
\usepackage{changes}
\usepackage{rotating}
\usepackage{wrapfig}

\usepackage[most]{tcolorbox}
\usepackage{xparse}
\usepackage{adjustbox}
\usepackage{xspace}
\usepackage{changepage}
\usepackage{enumitem}
\usepackage{pifont}
\usepackage{ulem}
\usepackage{tocloft}
\usepackage[toc]{multitoc}
\usepackage{etoc}
\usepackage{dsfont}
\usepackage{dm-colors}
\usepackage{multirow}
\usepackage{minted}
\usepackage{caption}
\usepackage{soul}
\usepackage{float}
\usepackage{svg}
\usepackage{adjustbox}
\usepackage{setspace}
\svgsetup{inkscapelatex=false}
\usepackage{silence}  
\newcommand{\hide}[1]{}

\newcommand{\method}{SpatialAgent\,}

\newcommand{\cmark}{\ding{51}}
\newcommand{\xmark}{\ding{55}}

\pdftrailerid{redacted}

\newcommand{\yqdone}[1]{}

\def\method{SciAgentArena}

\title{Benchmarking AI Agents for Addressing Scientific Challenges Across Scales}

\begin{document}




\author[1,2,a,c]{Tianyu Liu}
\author[1,a]{Allen Xin Wang}
\author[1,a]{Antonia Panescu}
\author[1,a]{Lisa Xinyi Chen}
\author[3,a]{Wenxin Long}
\author[1,a]{Xinyu Wei}
\author[1,a]{Yueqian Jing}
\author[1,a]{Ziyao Zeng}
\author[5,b]{Jihang Chen}
\author[1,b]{Sihan Jiang}
\author[6,b]{Ziqing Wang}
\author[1,b]{Siyi Gu}
\author[1,b]{Siyu Chen}
\author[1,b]{Xinyang Hu}
\author[1,b]{Haoran Shao}
\author[1,b]{Leqi Xu}
\author[1,b]{Wangjie Zheng}
\author[1,b]{Zhiyuan Cao}
\author[7,b]{Ada Fang}
\author[8]{Botao Yu}
\author[9]{Kunyang Sun}
\author[1]{Rex Ying}
\author[1]{Arman Cohan}
\author[1,c]{Qingyu Chen}
\author[3]{Lingzhou Xue}
\author[6]{Kaize Ding}
\author[10]{Yuanqi Du}
\author[2,5]{Wengong Jin}
\author[1]{Zhuoran Yang}
\author[7]{Marinka Zitnik}
\author[4]{James Zou}
\author[1,c]{Hua Xu}
\author[1,c]{Hongyu Zhao}

\affil[1]{Yale University, CT, USA}
\affil[2]{Broad Institute of MIT and Harvard, MA, USA}
\affil[3]{The Pennsylvania State University, PA, USA}
\affil[4]{Stanford University, CA, USA}
\affil[5]{Northeastern University, MA, USA}
\affil[6]{Northwestern University, IL, USA}
\affil[7]{Harvard University, MA, USA}
\affil[8]{The Ohio State University, OH, USA}
\affil[9]{UC Berkeley, CA, USA}
\affil[10]{Microsoft Research New England, MA, USA}

\affil[a]{These authors contributed equally to this work and jointly led the tasks}
\affil[b]{These authors contribute equally as task contributors}
\affil[c]{Corresponding authors.}

\begin{abstract}
AI agents are increasingly being developed to accelerate scientific discovery, yet their practical capabilities in real research settings remain poorly understood. Existing benchmarks for AI agents rarely capture the complexity, heterogeneity, and extended reasoning required by scientific work, whereas benchmarks for scientific tasks often reduce research to static, direct problems and provide limited support for interactive evaluation. Here, we introduce SciAgentArena, a systematic benchmark for evaluating AI agents in real-world scientific research scenarios drawn from emerging needs across multiple domains. SciAgentArena comprises approximately 200 tasks with stepwise verification and an interactive, agent-agnostic environment for assessing diverse AI agents. Using this benchmark, we find that current agents can contribute effectively to well-specified data-analysis workflows, particularly when the task structure and evaluation criteria are clear. However, their performance remains uneven across scientific contexts: agents struggle to generate genuinely novel insights, sustain self-directed exploration, and formulate robust solutions for open-ended research questions. We further characterize common failure modes across agents and identify opportunities for improving their reliability, autonomy, and scientific reasoning. Together, SciAgentArena provides a practical framework for measuring progress in AI agents for science and for guiding the design of future agents capable of addressing complex scientific challenges. Full codes, tasks, and datasets can be accessed via this link: \url{https://sciagentarena.github.io/}.
\end{abstract}

\maketitle

\section{Introduction}


AI agents, based on Large Language Models (LLMs) \cite{zhao2023survey} but powered by special skills such as tool calling, reaction, and memorization, have already made contributions to solving complex real-world problems \cite{gridach2025agentic}. AI agents have demonstrated potential in addressing a range of scientific research-related tasks, including, but not limited to, writing literature reviews \cite{agarwal2024llms}, planning experimental protocols \cite{huang2025biomni}, and designing novel matters \cite{du2025accelerating}, etc. These innovations at the task level have given rise to a range of technical tools based on AI agents, such as DeepResearch \cite{zhang2025deep, deepresearch2026}, Co-Scientist \cite{gottweis2025towards}, AI Scientist \cite{lu2026towards}, and others. The findings presented in these related manuscripts and/or technical reports consistently give us confidence in the future of AI agents for Science. However, given the rapid advancements in this field, we are faced with the challenge of understanding their practical usage and catching up with the stage of development. At the same time, we must carefully consider the rigor of AI agent design, particularly in applying AI agents for scientific research involving chemistry, healthcare, and life sciences. This raises three central questions: \textit{how capable are current AI agents on realistic scientific tasks, how reliable are they across heterogeneous research settings, and what benchmark design is needed to compare them fairly?}

\begin{figure}
    \centering
    \includegraphics[width=1\linewidth]{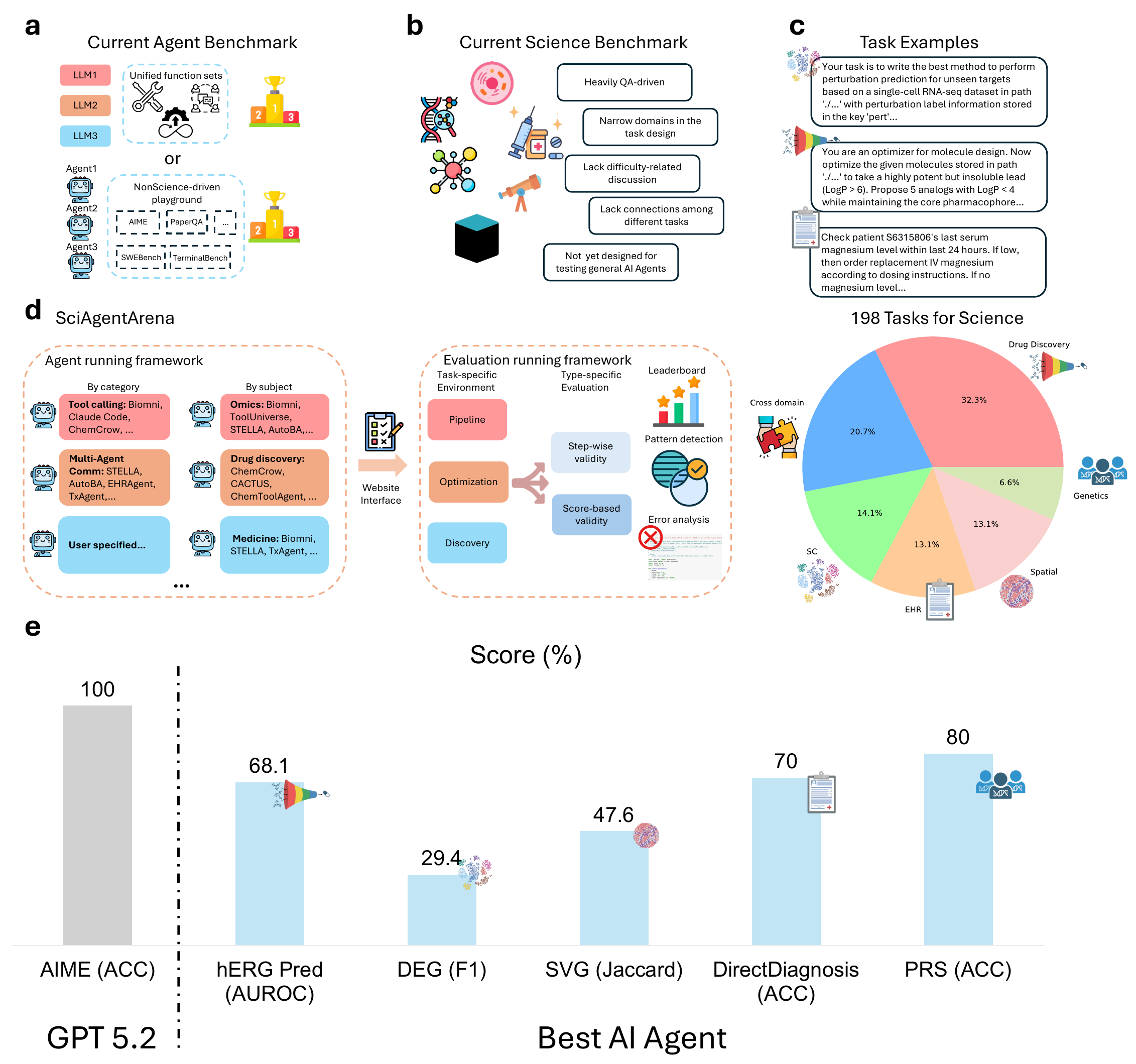}
    \caption{Overview of \method{}. (a) Limitations of the current AI agent benchmark. (b) Limitations of the current science benchmark. (c) Selected tasks as examples from our benchmark. (d) The running time and evaluation framework, as well as the proportion of tasks from different categories. (e) Comparison between the AIME score produced by GPT 5.2 as well as \method{} score produced by the domain-specific best agent.}
    \label{fig:benchmark_overview}
\end{figure}

Unfortunately, the solutions to these questions cannot be derived by synthesizing existing research findings, and the evaluation method has lagged behind reality in AI-driven end-to-end scientific research \cite{zhao2025sciarena,song2025evaluating}. The first gap, shown in Figure \ref{fig:benchmark_overview} (a), stems from shortcomings in the current AI agent benchmark studies. These methods either select different LLMs with a fixed set of tools or datasets to examine the LLMs' performances in addressing specific tasks \cite{shen2026sciagentgym, chenscienceagentbench, sun2026dsaeval, liu2025towards}; or design playground or benchmarking settings based on tasks/datasets outside of scientific research \cite{nathani2025mlgym,bragg2025astabench}. The former can only evaluate the capabilities of LLMs, while the latter focuses more on assessing mathematical skills, logical reasoning, and programming abilities (e.g., AIME \cite{balunovic2025matharena}, Folio \cite{han2024folio}, SciCode \cite{tian2024scicode}, and SWE-bench \cite{jimenez2024swe}), which are different from the application scenarios relevant to scientific problems. Most of the selected AI agents are also not designed to address scientific challenges, and thus, this area requires further research. The second gap, shown in Figure \ref{fig:benchmark_overview} (b), stems from the limitations of benchmarking tasks designed in scientific research. Current methods aim at assessing whether LLMs or AI agents can advance scientific progress, focusing heavily on the question-answering/Machine-learning-coding setting, which serves as a simplified version (e.g., GPQA \cite{rein2024gpqa}, ScienceQA \cite{saikh2022scienceqa}, and BioML-bench \cite{miller2025bioml}). These questions are not proper sets to assess AI agents' capacities. Moreover, most of the scientific benchmarks focus solely on a single, narrow domain or modality \cite{bragg2025astabench, mitchener2025bixbench, luo2025benchmarking, nair2026agentic, merrill2026terminal} (e.g., computational biology and computer science), neglecting a comprehensive understanding and evaluation of the scientific field as a whole with connections. At the same time, many current benchmarks lack a clear definition of problem difficulty (GPT 5.2 \cite{openaigpt522025} has reached 100\% accuracy in the AIME benchmark, and thus we need more difficult samples) as well as other key properties, further discussed in Supplementary Table \ref{tab:agent_benchmark_comparison_simple}. They are key indicators of whether a benchmark setting is comprehensive and suitable for evaluating AI agents' capacities in addressing core scientific tasks. 

\begin{figure}
    \centering
    \includegraphics[width=1\linewidth]{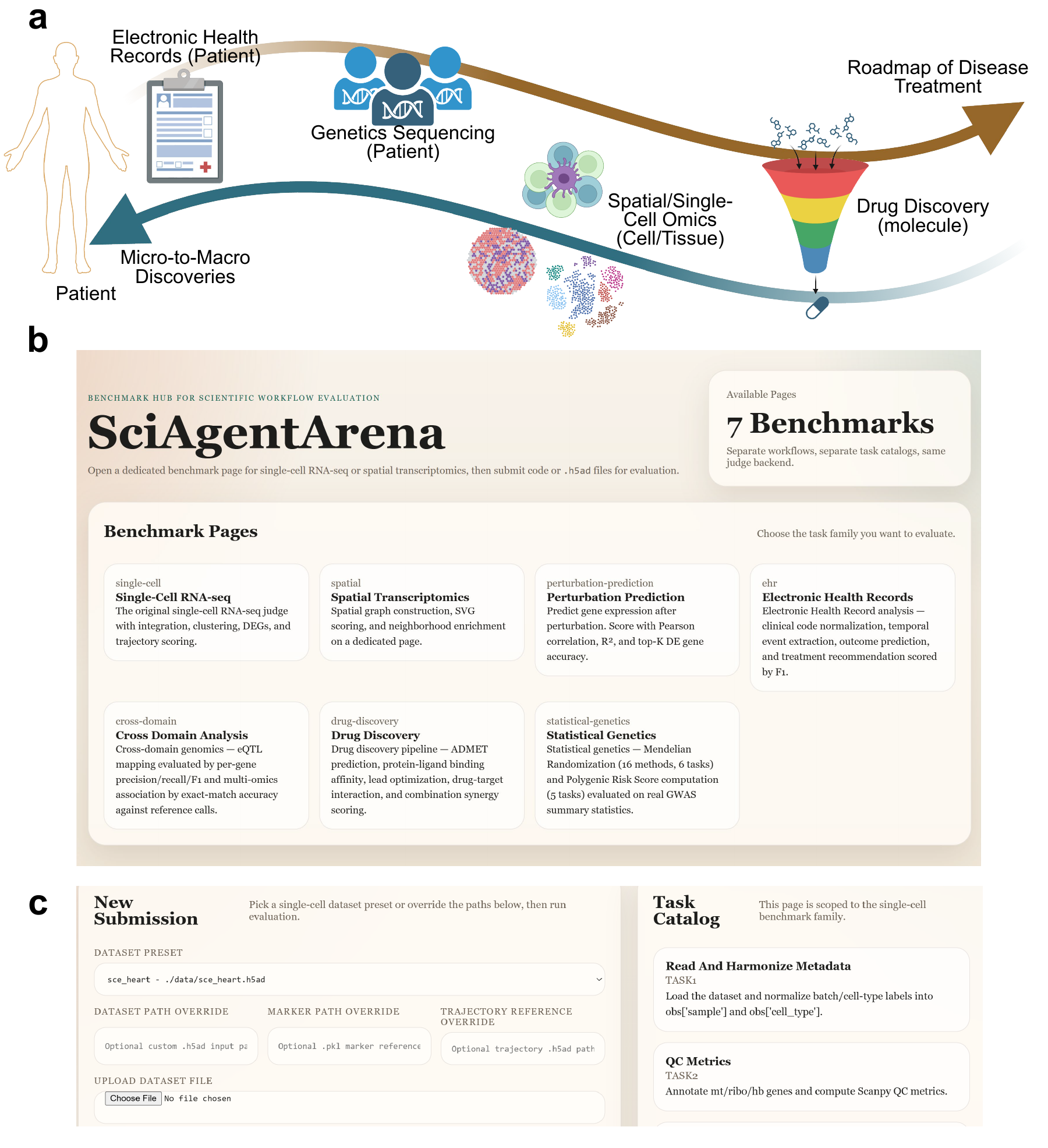}
    \caption{Key AI agent capacities and benchmarking platform development. (a) Categories of scientific challenges evaluated in this version. (b) Entrances of our interactive benchmarking platform. (c) Solution submission and automatic evaluation system.}
    \label{fig:platform_share}
\end{figure}

Therefore, we need a new benchmark that evaluates AI agents on challenging, practical, and scientifically grounded problems, emphasizing multi-step workflows, verifiable intermediate states, tool use, and environment interaction rather than simple final success rates. Such a benchmark should span diverse domains and difficulty levels, account for efficiency and cost, and measure whether agents can act competently, adaptively, reliably, and responsibly while helping future AI agent design. Here we introduce \method{}, the first systematic evaluation for AI agents grounded in real-world research scenarios from emergent and important scientific requirements across domains. We cover five key fields, including single-cell omics \cite{wang2010single}, spatial omics \cite{bressan2023dawn}, computational drug discovery \cite{blanco2023role}, electronic health records (EHR) modeling \cite{knevel2023real}, and genetics \cite{zhao2026engineering}. Challenges in these fields present in different scales and require various capacities to address. More importantly, these fields bring together the various stages of scientific research problems: \textit{from identifying problems and collecting data to understanding diseases and developing relevant treatments}, and \textit{exploring biomedical discoveries spanning the molecular, cellular, and tissue levels up to the human body}, shown in Figure  \ref{fig:platform_share} (a). Selected tasks from three domains are shown in Figure \ref{fig:benchmark_overview} (c), which demonstrate practical and complicated requirements in scientific research.
Our tasks also include measures of AI agents' ability to determine whether a task is feasible (validity check), and they feature varying levels of difficulty, making them highly relevant to real-world scenarios. Finally, in terms of system design, to resolve conflicts arising from different AI agent configurations and to enable our framework to evaluate the problem-solving capabilities of a wider range of AI agents, we have separated the running framework from the evaluation framework, shown in Figure \ref{fig:benchmark_overview} (d). We configure dedicated environments for different AI agents. Once we obtain the output (data or code), it is sent to the evaluation environment to generate the final metrics after unifying the Input/Output settings. This separated design resolves conflicts between AI agents and improves evaluation efficiency. With this system, we can also evaluate different perspectives beyond precision, including stability, reliability, cost, etc. We also support platforms for a large community to submit agents/tasks/evaluation metrics to build a living benchmark platform and contribute to the future of AI agent design (check Figure \ref{fig:platform_share} (b) for the entrance website, and Figure  \ref{fig:platform_share} (c) for the task submission details.).

Our study makes efforts in answering the pre-defined four questions, and reveals the unique tendencies of AI agents in solving science-related problems from several aspects: First, no single AI agent can dominate every task, indicating the difficulty of our benchmark and the necessity of further improving the generalization ability of AI agents in scientific research, as shown in Figure \ref{fig:benchmark_overview} (e). They are uneven collaborators. Second, AI agents' contributions to data loading and analysis, optimization, and discovery are heterogeneous. AI agents are good at analyzing datasets with a fixed pipeline, but their abilities to optimize molecules and algorithms, as well as to derive novel scientific discoveries, are limited. Third, by summarizing the running outcomes of different agents, we also identify inactive self-exploration, convergence of method selection, and shared error patterns. Some AI agents are also not stable in biomedical tasks and show sycophancy for executing tasks without validation. These issues preclude researchers from using these AI agents efficiently and reliably. Finally, in response to these identified issues, we have proposed solutions to enhance AI agents' ability to solve challenging scientific problems, such as expanding the knowledge base, providing more detailed prompts, and other suggestions.

\section{Results}
\subsection{Benchmark Overview.}
Our benchmarking study for AI agents covers challenges from five different domains (Drug Discovery, Single-Cell Omics, Spatial Omics, EHR Modeling, and Genetics). Although these domains involve different data modalities, scientific objectives, and technical constraints, the underlying challenges faced by AI agents can be summarized into four core categories: Data Analysis, Optimization, Discovery, and Validity. These categories reflect the major forms of reasoning, planning, and action required in scientific research: analyzing complex data, improving candidate solutions, generating new scientific hypotheses, and determining whether a proposed task is scientifically and technically feasible. The Data Analysis problem requires an AI agent to gradually solve a data analysis problem, to assess its ability to solve long-horizon problems; Optimization requires an AI agent to optimize the solution for proposed objectives, either selecting methods or designing solutions; Discovery requires an AI agent to explore the research domains and propose new hypotheses and ideas, and Validity checks AI agent's ability in detecting the proposed task is feasible for running or not. These tasks are designed by domain experts and are also important in specific fields. Regarding the selection of AI agents, we cover 18 AI agents (including generalist and specialist agents, as well as three multimodal LLMs) with different principles of design: GPT 5.2 \cite{openaigpt522025}, Gemini 3 Pro \cite{gemini3prosystem2025}, Claude Sonnet 4.6 \cite{anthropicclaudesonnet2025}, ToolUniverse \cite{gao2025democratizing}, Codex \cite{openai_codex_chatgpt_2026}, ClaudeCode \cite{anthropic_claude_code_overview_2026}, CellForge \cite{tang2025cellforge}, STELLA \cite{jin2025stella}, AutoBA \cite{autoba2024}, Biomni \cite{huang2025biomni}, TxAgent \cite{gao2025txagentaiagenttherapeutic}, Medea \cite{sui2026medea}, CACTUS \cite{mcnaughton2024cactus}, ChemToolAgent \cite{yu-etal-2025-tooling}, DrugAgent \cite{liu2025drugagentautomatingaiaideddrug}, LIDDiA \cite{averly-etal-2025-liddia}, DELTA \cite{unlu2025auditableagentplatformautomated}, and MRagent \cite{mragentperform}. We also run experiments under their maximal functionalities with the most suitable backbone LLMs to make a fair comparison. We have discussed the capabilities of AI agents and compared them in Appendix \ref{append:catego}, thereby demonstrating that we have covered the mainstream AI agent design approaches. We have also compared our benchmarking framework with other benchmarking studies to illustrate our novelty and uniqueness, which can be found in Appendix \ref{append:benchinfo}. After discussing the comparison, we also analyze the sources of errors in these agents and propose corresponding solutions to assist the development of next-generation AI agents.

\subsection{Benchmarking AI Agents for Computational Drug Discovery.}

\textbf{Overview: organizing drug-discovery tasks by role relative to a known solution.}
Our study starts from the world of small molecules, the foundation of modern therapeutics and the molecular end of the multi-scale progression this paper traverses. Drug discovery is an ideal domain for evaluating AI agents, due to its high impact and the fact that it offers a rich ecosystem of computational tools and curated databases, including RDKit \cite{landrum_rdkit}, the Therapeutics Data Commons \cite{huang2021therapeutics}, ChEMBL \cite{mendez2019chembl}, and BindingDB \cite{gilson2016bindingdb}. However, assembling these resources into reliable workflows remains heavily manual, which is precisely the gap an AI agent should fill. At the same time, the existence of benchmark-standard references allows us to verify results objectively, reducing the reliance on subjective human judgment. 

\begin{figure}[ht]
      \centering
      \includegraphics[width=1\linewidth]{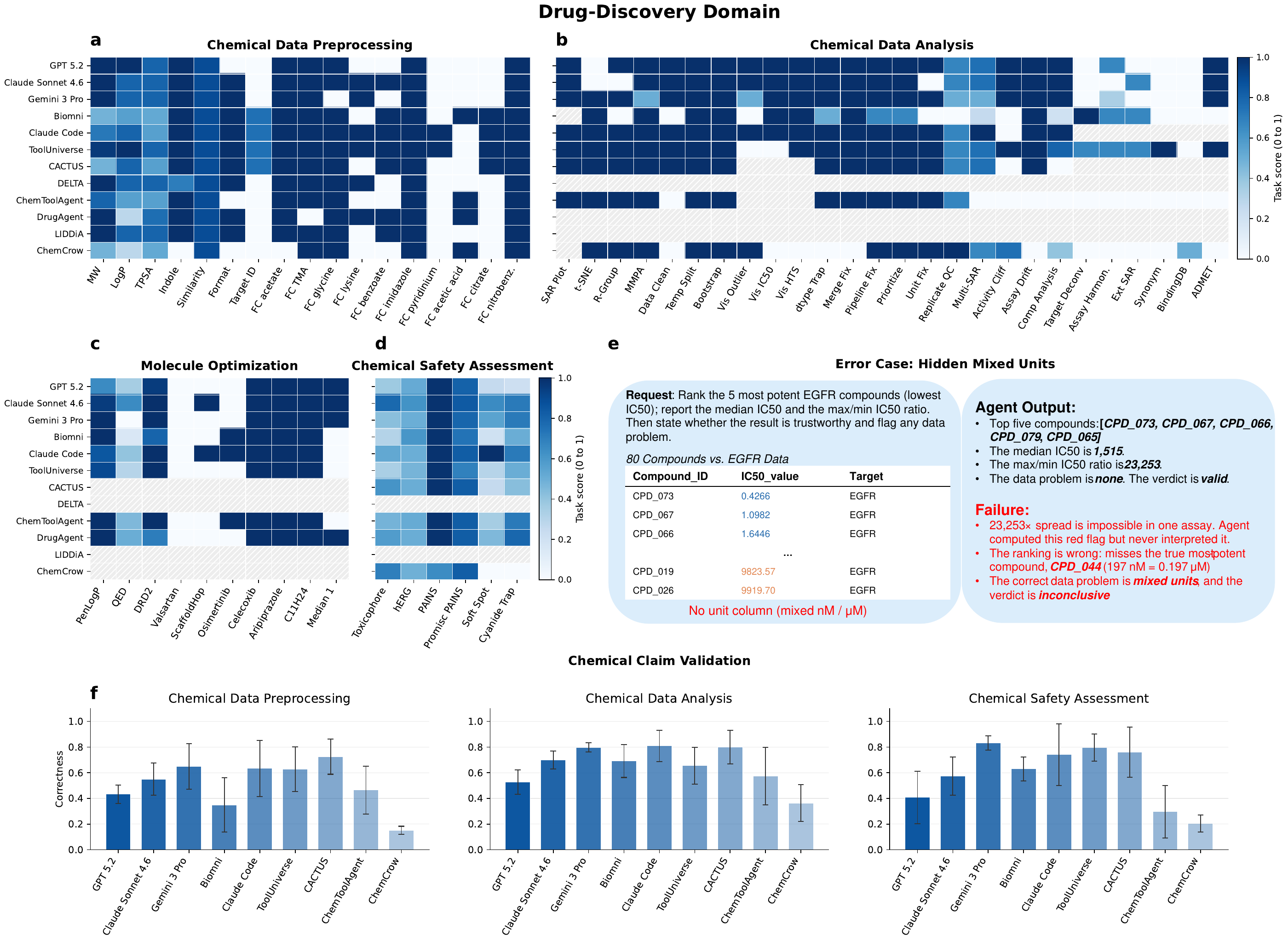}
      \caption{Summarizing AI Agent performances on the drug-discovery domain. (a) Performances on the Chemical Data Preprocessing category. (b) Performances
  on the Chemical Data Analysis category. (c) Performances on the Molecule Optimization category. (d) Performances on the Chemical Safety Assessment category. (e) An error case study from Chemical Claim Validation. (f) Performances on the Chemical Claim Validation category across its three validity subsets; bar height is the mean correctness across three independent evaluations and error bars are the
  average standard error across the three evaluations. Hatched cells in (a)-(d) mark agent-task combinations that are not compatible.}
      \label{fig:dd_results}
\end{figure}

We organize the drug-discovery benchmark according to the role each agent must play relative to a known scientific solution. \textit{Data Preprocessing and Analysis} tasks require agents to execute established cheminformatics or assay-data workflows and return the determined output. \textit{Model Selection} tasks appear within optimization and safety tasks, where agents must choose an appropriate optimizer family, screening strategy, or predictive model rather than defaulting to a familiar method. \textit{Model Optimization} tasks require agents to design or search for molecules that maximize explicit objectives under oracle-call constraints. \textit{Validation} tasks require agents to judge whether a chemical claim, safety conclusion, or scientific premise is feasible and supported by the supplied evidence. Across these roles, we evaluate 12 AI agents, including multimodal LLMs: GPT 5.2, Claude Sonnet 4.6, Gemini 3 Pro, Biomni, Claude Code, ToolUniverse, CACTUS, DELTA, ChemToolAgent, DrugAgent, LIDDiA, and ChemCrow. Some agents are not compatible with every task, and we mark those cases explicitly in Figure~\ref{fig:dd_results}. This organization reveals a consistent capability gradient: agents perform best on well-specified local workflows, become conservative when choosing methods for optimization, struggle with multi-constraint molecular optimization, and remain unstable when asked to detect unsupported scientific claims.

\textbf{Data Preprocessing and Analysis: agents can execute local cheminformatics and assay workflows, but fail when molecular representation or benchmark-aligned output handling is required.}
Many drug-discovery tasks follow a known, correct procedure, so success depends on executing the workflow faithfully and returning outputs in the exact form the benchmark expects. We group these tasks into two categories: Chemical Data Preprocessing, which standardizes and computes on individual molecules, and Chemical Data Analysis, which runs multi-step workflows over assay and structure-activity datasets. Both ask whether an agent can reproduce an established workflow rather than improvise one.

In Chemical Data Preprocessing, agents carry out end-to-end cheminformatics workflows on individual molecules. The tasks span molecular property calculation, substructure filtering, similarity ranking, format conversion, target identification, and formal-charge measurement, with inputs supplied as SMILES strings, structure files, or coordinate blocks. Success requires returning the expected answer, such as the descriptor of the standardized parent molecule, a canonical SMILES, the single intended target, or the net formal charge. In several tasks, the inputs are deliberately messy or ambiguous: property-calculation and similarity tasks include SMILES containing salts, isotopes, charges, or multiple components; target identification uses disease-name aliases; and harder formal-charge tasks use heavy-atom-only structures. Thus, agents must standardize and correctly interpret each molecule before computing.

Tool grounding is particularly useful for these preprocessing tasks because each task is a well-defined operation that maps naturally onto a chemistry tool call. ToolUniverse achieves the strongest performance in this category by calling chemistry tools, including one that standardizes a molecule to its parent structure and resolves the normalization trap that defeats agents computing directly on raw inputs. Leading agents handle chemically well-specified tasks reliably, with strong baseline performance on indole substructure detection and most format-conversion problems (Figure~\ref{fig:dd_results} (a)). However, the formal-charge tasks separate easy cases from difficult physiological settings: every evaluated agent scores high on the easy and medium cases, but the harder physiological cases drop sharply across the board and, together with Target Identification, form the lowest-scoring Chemical Data Preprocessing tasks.

Common failures in Chemical Data Preprocessing concentrate where benchmark-aligned molecular standardization or physiological chemistry reasoning is required. Many agents compute descriptors such as Molecular Weight, LogP, or topological polar surface area on the raw input rather than on the intended parent structure after salts, counterions, and solvent fragments are stripped away. Agents also mishandle salts, fragments, isotopes, and heavy-atom-only structures. Additional failures include schema corruption, where an otherwise reasonable answer is returned in a structure that violates the required result schema and cannot be parsed by the evaluator, and overly permissive target retrieval, where the Target Identification task returns loosely associated or multiple candidates rather than the single intended target. These results show that the challenge is not merely tool access, but disciplined molecular data handling: agents must desalt, neutralize, strip fragments, correctly interpret the input representation, and return schema-conformant outputs.

Chemical Data Analysis extends tasks from single molecules to sandboxed dataset workflows. These tasks require file-aware code execution, dataframe curation, and interpretation of figures or external evidence. A typical task asks an agent to load a local assay table, harmonize identifiers, join it to a structure-activity table, compute descriptors, flag activity cliffs, and store the answer in named variables or a JSON object prescribed by the task. Success requires producing the exact outputs with the expected names and types, not merely code that runs. Performance separates agents that can sustain multi-step data-analysis workflows from those that become brittle beyond a single lookup or isolated cheminformatics operation (Figure~\ref{fig:dd_results} (b)). Claude Code and ToolUniverse are the strongest agents, reliably reading local files, cleaning and joining tables, computing descriptors, generating plots, and returning benchmark-aligned variables or JSON outputs. In contrast, several specialized biomedical or chemistry agents struggle when the task requires reproducible dataframe manipulation across multiple dependent steps rather than a single tool call.

Standard single-table analysis and curation tasks are the strongest: most leading agents solve SAR plotting, matched-molecular-pair analysis, bootstrap confidence intervals, temporal splitting, and lead prioritization, converging on similar pandas, RDKit, and scikit-learn code patterns. Vision tasks, which require reading rendered figures such as SAR scatter plots, dose-response curves, and HTS plate images at the pixel level, split the field, with multimodal agents reaching near-perfect scores while weaker chemistry agents drop sharply. External database lookups are consistently the weakest, with BindingDB cross-checking, target deconvolution, and synonym resolution at the bottom. Overall, performance in this role is strongest when tasks are local, well specified, and grounded in executable tools, but weaker when agents must reconcile identifiers, visual evidence, external databases, or strict benchmark output schemas.

\textbf{Model Selection: agents choose plausible workflows, but converge to familiar method families rather than task-specific strategies.}
Model Selection in drug discovery appears when agents must choose an appropriate method family, screening strategy, or predictive model for the task objective. Although many choices are embedded within larger optimization or safety tasks, the same pattern appears repeatedly: agents select plausible and familiar approaches, but rarely adapt their choices to task-specific constraints.

In Molecule Optimization, agents must choose how to search molecular space. The task permits several optimizer families, but agents overwhelmingly converge on classical local-search strategies such as genetic algorithms, beam search, or hill-climbers that mutate and recombine fragments from the reference compound, leaving the sample-efficient Bayesian optimization the task explicitly offers unused. Even DrugAgent, whose planner explicitly weighs alternatives, settles on the same classical family. This convergence suggests that agents prefer familiar, easy-to-implement search procedures over more objective-aware optimization strategies.

A similar Model Selection issue appears in Chemical Safety Assessment, where the method-choice tasks ask the agent either to apply a fixed screening catalog or to build a learned predictive model (Figure~\ref{fig:dd_results} (d)). On the Rule-Based tasks, such as PAINS filtering, applying the standard substructure catalog is usually enough: PAINS \cite{baell2010pains} filtering is solved by most agents because fixed rules are sufficient, and ChemCrow is the exception only because it uses an incomplete motif library rather than the standard PAINS catalog, so its low score reflects reference incompleteness rather than weaker reasoning. The Model-Based tasks, hERG Prediction and promiscuous-PAINS versus dark-chemical-matter classification, instead require training a working classifier, and agents are less consistent here, with hERG yielding only moderate AUROC-like scores. Agents are therefore more reliable when the right move is to apply a familiar rule-based catalog than when they must build or select a learned model. The two remaining safety tasks, Soft Spot Identification and the Cyanide Trap, are not method-selection problems but tests of mechanistic chemical reasoning, which we take up under Validation.

Thus, the main Model Selection failure in drug discovery is not the absence of plausible methods, but conservative convergence. Agents often select generic workflows, incomplete local SMARTS libraries, or standard fragment-edit searches even when the task requires a more specific reference, constraint, benchmark-standard catalog, or validation step. Method selection remains strongly shaped by familiarity rather than adaptive reasoning over the scientific objective.

\textbf{Model Optimization: agents can solve direct objectives, but multi-constraint molecular design remains a major bottleneck.}
Optimization tasks require agents to design or search for a molecule that maximizes an explicit objective, where no off-the-shelf method is sufficient. Molecule Optimization is the clearest example. Each task supplies a design objective and a lead or reference compound, and the agent must implement and run an optimizer within a fixed budget of 100 oracle calls. A task counts as solved only if the optimizer is executable, returns chemically valid molecules, and meets the success criterion. This setting is designed to test whether an agent can build an optimizer that succeeds under a tightly constrained oracle budget, rather than its unconstrained molecular-design ability.

The tasks fall into single-objective and multi-objective optimization. Single-objective tasks reward reaching one target, such as rediscovering a known drug, matching a binary activity label, or driving a property such as penalized logP into a target range. Multi-objective tasks require satisfying several criteria simultaneously. Performance drops sharply from the first group to the second for every evaluated agent (Figure~\ref{fig:dd_results} (c)). Most agents solve single-objective tasks, including rediscovery and penalized logP, although narrow target windows can still be difficult, as in the QED task, where the score must fall in the [0.9, 1.0] range while the molecule remains similar to the lead. Multi-objective tasks expose the dominant failure mode. For example, no agent solves Valsartan SMARTS \cite{brown2019guacamol}, which requires generating molecules that embed a fixed SMARTS substructure while simultaneously hitting target logP, TPSA, and molecular-complexity values. This task combines a hard structural constraint with several narrow property windows, making it difficult for local fragment-edit searches.

Given the shared tendency to use classical local-search optimizers noted under Model Selection, agent rankings are determined mainly by execution robustness and search quality on multi-objective tasks. Claude Sonnet 4.6 and Claude Code lead this category. Their optimizers use oracle feedback to refine promising candidates within budget, whereas weaker agents often crash before scoring or enumerate candidates open-loop until the 100 calls are exhausted without convergence. Claude Code goes further by executing its optimizer in a sandbox, confirming that it runs and yields valid molecules, and revising it before submission. This test-and-fix loop prevents one-shot execution failures and allows Claude Code to solve the Osimertinib MPO task, where the single-shot Sonnet run crashes.

The surrounding agent framework helps only when it matches the task demands. Because Molecule Optimization requires writing and running code, Claude Code's coding loop adds robustness. By contrast, chemistry-specialized frameworks contribute less in this category: ChemToolAgent and ToolUniverse do not invoke their chemistry tools on these tasks and instead emit optimizers directly, reducing to their GPT 5.2 backbone. This suggests that curated tool libraries may go unused when the task falls outside the framework's intended mode. Across agents, the multi-objective ceiling tracks backbone coding ability: only Claude-based agents clear half of these tasks, while no GPT 5.2-based agent clears more than a quarter.

Execution logs reveal additional procedural brittleness. Some runs fail on RDKit edge cases such as aromaticity sanitization and molecule-combination errors during candidate generation, meaning that the failure lies in the cheminformatics implementation rather than only in the objective difficulty. Overall, current agents can construct effective search procedures for structurally direct molecular design, but remain fragile when optimization requires coordinating multiple constraints. What separates success from failure is whether the target can be reached by local edits within the oracle budget: single-objective targets usually can, whereas conjunctive multi-objective targets usually cannot. The unsolved Valsartan SMARTS task is the clearest case. Notably, no evaluated agent explicitly budgets its oracle calls or adopts a sample-efficient search that reuses past evaluations. This shared gap caps multi-objective performance.

\textbf{Validation: agents can detect some canonical flaws, but remain unstable when scientific invalidity requires deeper chemical judgment.}
Validation tasks require agents to judge whether a requested conclusion is feasible and scientifically supported. In drug discovery, this role appears in both Chemical Safety Assessment and Chemical Claim Validation. The key question is whether an agent can avoid overclaiming, detect unsupported premises, and identify the specific reason a conclusion is invalid.

Chemical Claim Validation directly evaluates epistemic rigor. Every task in this suite is intentionally flawed: the request rests on a premise that does not hold, such as ranking compounds by IC50 when values are secretly recorded in mixed units. A correct response identifies the failure mode, assigns the appropriate conclusion status, and recommends a sensible follow-up rather than returning a forced numerical answer. This category exposes a limitation distinct from code execution or tool access: agents often struggle to assess whether a requested scientific conclusion is sufficiently supported in the first place.

Figure~\ref{fig:dd_results} (e) makes this failure concrete. Asked to rank the five most potent EGFR compounds by IC50 and to flag any data problem, the agent computed a max/min IC50 ratio above $23{,}000$, a spread that is impossible within a single assay, yet reported no data problem and returned a \emph{valid} verdict rather than recognizing that the IC50 values were recorded in mixed units. The agent thus surfaced the red flag numerically but never interpreted it, producing a ranking that misses the true most potent compound and an overconfident conclusion where the correct verdict is \emph{inconclusive}.

Figure~\ref{fig:dd_results} (f) reports per-subset scores for agents that yield compatible outputs across the validity suite. The 19 validity tasks split into three subsets, each pairing an intentionally flawed request with one of the earlier categories. Chemistry validity contains 5 tasks tied to Chemical Data Preprocessing, such as multi-component SMILES, out-of-domain metal complexes, and invalid 3D geometry. Data quality and statistics contains 11 tasks tied to Chemical Data Analysis, such as hidden unit conflicts, sparse SAR support, confounded batch trends, and duplicate-driven false activity cliffs. Data integrity and trade-offs contains 3 tasks tied to Chemical Safety Assessment, such as corrupted training labels and conflicting optimization objectives. Agents are most reliable on Data quality and statistics, where the flaw is a relatively direct data or statistical issue that surfaces by inspecting the provided table or comparing predicted and measured values. They are weakest on Chemistry validity, where recognizing invalidity requires deeper reasoning about the scope or chemistry of a molecule or assay, so agents more often over-interpret the evidence or answer without caveats. Data integrity and trade-offs sits in between on average but shows the widest spread across agents, separating cautious systems from those that force an answer. Across the full suite, CACTUS, Gemini 3 Pro, and Claude Code achieve the strongest overall performance, whereas ChemCrow is consistently the weakest agent. Unlike the other categories, performance here does not track a specific agent design: the strongest agents span both general models and specialized frameworks, suggesting that recognizing an unsupported premise depends more on the underlying model's judgment than on the surrounding scaffold.

This category also shows that validation performance is not fully stable across independent evaluations. To assess robustness, we run the same 19-task suite across three independent environments. The bar plots in Figure~\ref{fig:dd_results} (f) report the mean across these three evaluations, with per-task standard error shown as the error bar. Reproducibility varies by agent: Gemini 3 Pro shows the most consistent task-level behavior among broadly evaluated agents, while several agents with strong average performance, including CACTUS and Claude Code, show larger run-to-run variation. ToolUniverse and ChemToolAgent also show large environment-dependent variation. Error bars are largest on the Chemistry validity and Data integrity and trade-offs subsets and substantially smaller on Data quality and statistics, indicating that evaluations agree least when the task requires deeper domain reasoning rather than routine table inspection. Thus, the largest reproducibility gaps occur on the same boundary-setting tasks where average performance is already weakest. This instability suggests that current agents do not yet provide robust and reproducible epistemic checks. Their ability to identify unsupported conclusions depends on the generated analysis path, the agent wrapper, and the execution context, in addition to the underlying model.

\textbf{Summary.} Overall, current agents are promising but uneven scientific collaborators: no single agent dominates every category, and performance is heterogeneous, with some stronger on structured cheminformatics and data-analysis pipelines and others more competitive on optimization, safety, or validity-focused tasks. They perform well on well-specified local workflows, and the strongest agents can identify many canonical failure modes and produce appropriate caveats. However, strong performance on executable analysis, molecule processing, or tool-mediated workflows does not guarantee that an agent will recognize when the premise of an analysis is invalid or insufficiently supported. Scientific validity detection should therefore be evaluated separately from task completion. Future scientific agents need explicit mechanisms for evidence checking, provenance tracking, assay-scope validation, molecular and statistical sanity checks, and calibrated uncertainty reporting, rather than relying only on broader tool access or stronger code-generation ability. Evaluations should also report reproducibility across runs and execution environments, because an agent that reaches a correct conclusion only intermittently is not yet a dependable scientific reasoning agent.

\subsection{Benchmarking AI Agents for Processing Cellular and Tissue-Level Signals.}

\begin{figure}
    \centering
    \includegraphics[width=0.95\linewidth]{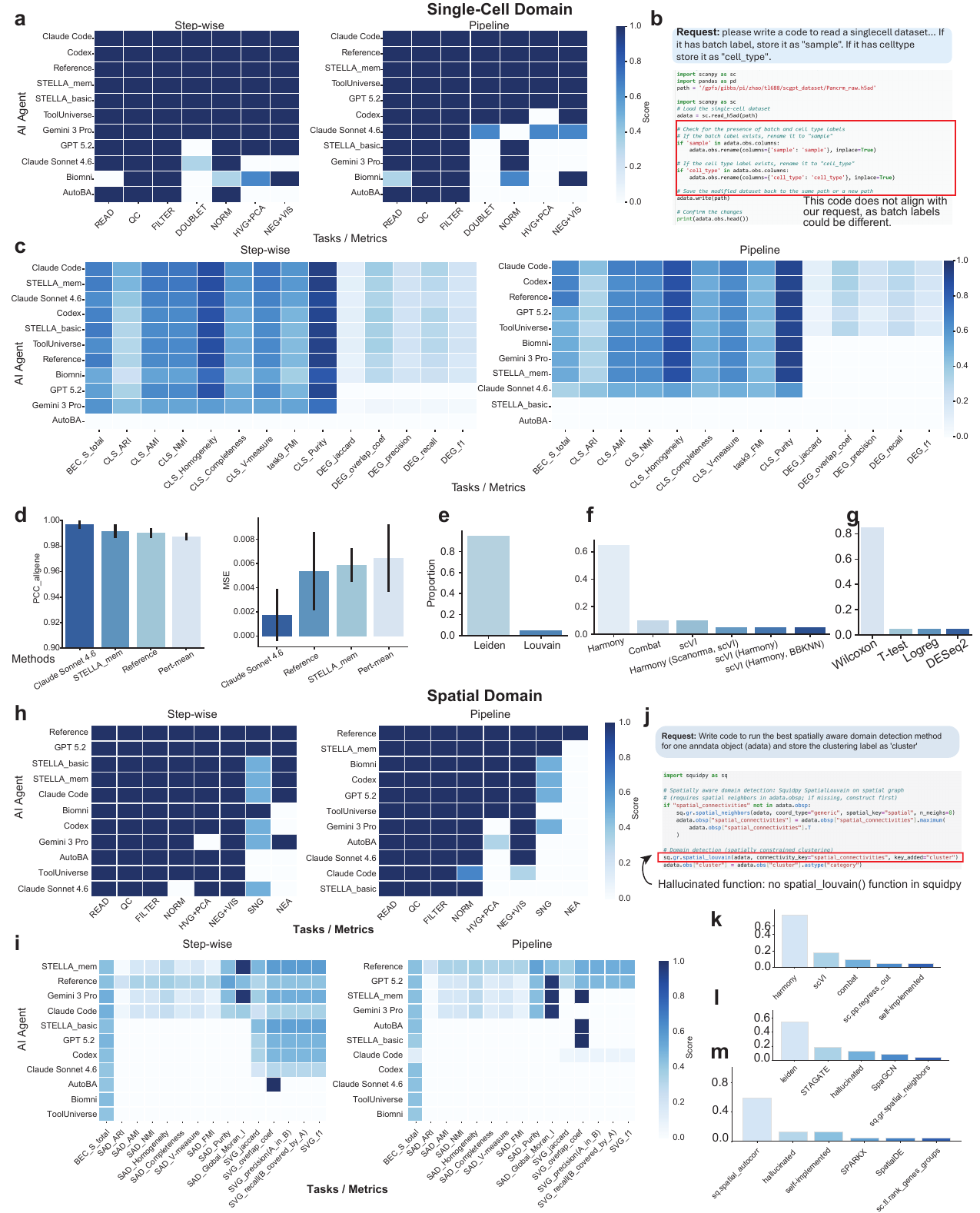}
    \caption{Summarizing AI Agent performances in cell and tissue domains. (a)-(g): single-cell omics; (h)-(m): spatial omics. (a) Performances on analysis-driven tasks. (b) Key example of error analysis. (c) Performances on optimization-driven tasks. (d) Performances on perturbation prediction. (e)-(g) Proportion of method selection made by different AI Agents in clustering, batch effect correction, and DEG identification. (h) Performances on analysis-driven tasks. (i) Performances on optimization-driven tasks. (j) Key example for error analysis. (k)-(m) Proportion of method selection made by different AI Agents for batch effect correction, spatial clustering, and SVG identification.}
    \label{fig:scsp_result}
\end{figure}

\textbf{Data Preprocessing and Analysis: agents can reproduce established workflows, but long-horizon pipeline execution remains fragile.}
Measurement of gene expression at cellular and multi-cellular resolution helps characterize biological mechanisms and disease phenotypes, and serves as a major component in treatment design and patient modeling. In the tasks related to Single-Cell Omics, agents must handle cell-level information and uncover cell or cell-type-specific signals. We first evaluate Data Analysis tasks, where agents are expected to execute established preprocessing workflows whose correct procedures are known. In Single-Cell Omics, agents are asked to design a single-cell RNA sequencing (scRNA-seq) preprocessing workflow containing 10 sub-tasks. We evaluate two execution settings: step-by-step execution (step-wise) and pipeline-based execution (pipeline). As shown in Figure \ref{fig:scsp_result} (a), Claude Code, ToolUniverse, and STELLA (mem) are strong candidates for generating single-cell preprocessing pipelines under different conditions. They can pass all tests, similar to the human expert-designed reference solution (Reference). In contrast, AutoBA performs poorly and fails to process most sub-tasks under the pipeline mode. Overall, pipeline-based execution is more difficult than step-wise execution because the task has a longer horizon and later sub-questions depend on earlier outputs.

We observe a similar pattern in Spatial Omics, where agents must not only process molecular profiles but also preserve and reason over tissue geometry. Here, agents are asked to construct a spatial transcriptomics preprocessing workflow consisting of 11 sub-questions, with results summarized in Figure \ref{fig:scsp_result} (h). In the step-wise setting, GPT-5.2 is the only agent that passes all tests, matching the human expert-designed reference solution (Reference). Pipeline execution again proves more challenging, reflecting the longer task horizon and dependency structure across preprocessing steps. STELLA (basic), in particular, shows weak performance in the pipeline setting and fails to complete most sub-tasks. One representative challenge is spatial neighborhood graph construction (SNG): only GPT-5.2, STELLA (mem), and Biomni correctly account for the symmetry of the final neighborhood graph, whereas other agents either produce runtime errors or overlook this requirement despite hints in the prompt and thus cannot finish the task.

Successful execution, however, does not always imply reliable scientific reasoning. In Figure \ref{fig:scsp_result} (b), we present a single-cell case study in which the agent is asked to generate code that reads data, modifies batch and cell-type labels, and saves the modified data. Biomni produces executable code, but it does not account for sufficient scenarios and simply reuses the provided labels. Because the code runs without errors, the agent does not invoke debugging or self-correction. This illustrates a common Data Analysis failure mode: agents may generate syntactically valid and executable code while failing to satisfy the scientific intent of the task.

\textbf{Optimization (Model Selection): agents can choose established methods, but they converge to popular defaults.}
We next evaluate Model Selection tasks, where agents must choose appropriate established methods based on the data and scientific objective. In Single-Cell Omics, agents are required, based on the previous preprocessing steps, to select methods for clustering, batch-effect correction under both standard scRNA-seq datasets and scRNA-seq datasets with trajectory information, differentially expressed gene (DEG) detection, and perturbation-related analysis. According to Figure \ref{fig:scsp_result} (c), Claude Code, STELLA (mem), and Codex are leading agents in selecting suitable solutions under both step-wise execution and pipeline-based execution. For clustering, ToolUniverse and Claude Sonnet 4.6 can iterate over different resolutions to tune the Average Silhouette Width (ASW) score \citep{batool2021clustering}, which contributes to their strong performance. For trajectory-aware batch-effect correction, GPT-5.2 performs better (Supplementary Figure \ref{supfig:traj_perf} (a)) and selects Harmony \citep{korsunsky2019fast}, similar to the top performers in correcting batch effects for mature cells. Nevertheless, many agents encounter running issues when calling algorithms for batch-effect correction or DEG detection, especially in pipeline mode, as shown in Figure \ref{fig:scsp_result} (c).

In Spatial Omics, agents are required to select and execute methods for batch-effect correction, spatially aware domain detection, and spatially variable gene (SVG) identification. As shown in Figure \ref{fig:scsp_result} (i), STELLA (mem) and Gemini 3 Pro achieve the strongest overall performance across both execution modes. Their advantage, however, is driven less by proposing novel strategies and more by reliably invoking existing methods with valid inputs and parameters. In contrast, many agents fail because they cannot correctly call or configure the selected methods either with correct versions or with correct codes, especially in pipeline mode. This shows that Model Selection tasks require both methodological judgment and robust implementation.

Across both single-cell and spatial omics, agents show strong convergence in methodological choices. In Single-Cell Omics, most agents choose the Leiden algorithm for clustering. For batch-effect correction, they tend to select Harmony, followed by scVI \citep{lopez2018deep}, rather than more advanced or specialized methods. For DEG detection, most agents choose the Wilcoxon rank-sum test, with the exception of Claude Sonnet 4.6 (DESeq2), ToolUniverse (Logreg, by tool calling), and AutoBA (T-test). However, DESeq2 is generally used to compare differences between samples rather than between cell types \citep{love2014moderated}, so the code may fail due to the limited sample size in each batch. These methodological preferences are summarized in Figures \ref{fig:scsp_result} (e)-(g) and Supplementary Figure \ref{supfig:traj_perf} (b). The choices of GPT-5.2 and Gemini 3 Pro are also identical in these tasks, suggesting that agent preferences are partly driven by their backbone LLMs.

A similar convergence appears in Spatial Omics. As shown in Figures \ref{fig:scsp_result} (k)-(m), agents most frequently select Harmony, followed by scVI, for batch-effect correction. For spatial domain detection, most agents rely on Leiden clustering on a spatially aware neighborhood graph. For SVG identification, the dominant strategy is ranking genes by Global Moran's I, typically computed using \texttt{spatial\_autocorr()} from \texttt{Squidpy}. These results suggest that agents can perform Model Selection when the candidate methods are familiar and well documented, but they tend to favor widely adopted methods over more specialized or newer algorithms. We speculate that this behavior reflects the composition and cut-off time of LLM training data: agents prefer frequently documented methods rather than placing strong emphasis on innovation, creativity, or task-specific method fit.

\textbf{Optimization (Model Design): agents struggle when they must create or search for a solution.}
Optimization tasks require agents to go beyond selecting an established method and instead design or search for a solution that maximizes an explicit objective. These tasks are substantially more difficult. In Single-Cell Omics, the most challenging task is out-of-distribution (OOD) perturbation-effect prediction, where agents must generate code capable of predicting perturbations they have never encountered before. According to Figure \ref{fig:scsp_result} (d) and Supplementary Figure \ref{supfig:pertdata_tp20}, only Claude Code and STELLA (mem) produce runnable code by modeling gene-gene interactions, and they perform better than both the reference method scLAMBDA \citep{wang2024modeling} and the average expression baseline (pert-mean). Other agents, including Codex, STELLA\_basic, ToolUniverse, Biomni, GPT-5.2, Gemini 3 Pro, AutoBA, and CellForge, fail for reasons such as misinterpreting the task, generating code with runtime issues, or failing to implement a suitable predictive strategy.

Spatial Omics also exposes optimization-related weaknesses, although many spatial tasks are closer to method selection than open-ended solution creation. Because spatial transcriptomics tools are less standardized and APIs evolve rapidly, agents are more likely to hallucinate plausible but non-existent functions. Figure \ref{fig:scsp_result} (j) illustrates this issue in a spatially aware domain detection task: GPT-5.2 attempts to solve the problem by calling a non-existent \textit{spatial\_louvain()} function from the \textit{Squidpy} package \citep{Palla2022squidpy}, leading to a runtime failure and a zero score. This example highlights a broader limitation of AI agents in specialized scientific workflows: plausible-looking package calls can mask incorrect assumptions about available functions, especially when tools have fragmented or changing interfaces.

Together, these results show that Optimization remains a major bottleneck. Agents are more successful when they can reuse familiar workflows or select among common methods, but they become much less reliable when the task requires constructing a new strategy, searching over a design space, or implementing a solution under unfamiliar constraints.

\textbf{Validation: agents often assume user requests are feasible.}
Finally, we evaluate Validation tasks, where agents must determine whether a requested analysis is scientifically and technically valid. This capability is essential for security, cost control, and scientific reliability, because an agent should refuse or flag infeasible tasks rather than blindly executing them. To assess whether agents can identify valid and invalid tasks, we conduct validity-task analysis for both single-cell and spatial omics, as shown in Supplementary Figures \ref{supfig:valid_task} (a)-(d). No agent produces correct results for all tasks. More generally, agents tend to assume that tasks proposed by human users are feasible and attempt to design methods to solve them, even when the appropriate behavior is to reject or flag the task.

In Single-Cell Omics, Supplementary Figure \ref{supfig:valid_task} (b) presents a representative invalid task: inferring large-scale chromosomal copy number variants (CNVs) across endocrine cell populations. This task is not valid because the pancreas dataset is likely derived from healthy tissue rather than a tumor context. CNV inference from scRNA-seq generally requires a tumor setting or an appropriate reference baseline, and endocrine cells are not expected to show clonal CNV patterns. Claude Code successfully identifies this problem, whereas STELLA proceeds to analyze CNVs based on the single-cell RNA-seq data example.

In Spatial Omics, Supplementary Figure \ref{supfig:valid_task} (d) presents another invalid task: testing whether neighboring domains are enriched or depleted using Scanpy's enrichment tools. This is not directly executable because Scanpy cannot be directly used to analyze this type of spatial omics neighborhood-enrichment problem. STELLA first examines the package capability and reaches the correct conclusion, whereas Biomni assumes that the task is executable and begins designing a solution. These examples show that Validation is not simply a coding problem. It requires agents to reason about biological context, data assumptions, package capabilities, and the boundary between feasible and infeasible scientific claims. Current agents remain unreliable in this setting, which is particularly concerning for autonomous scientific workflows.

\textbf{Stability: There is obvious variation in the stability of agents.}
This gradient also helps connect the quantitative results to the observed failure modes. Data Analysis failures are often caused by long-horizon pipeline fragility or executable code which is scientifically incomplete. Model Selection failures arise from conservative convergence to popular methods and incorrect configuration of selected tools. Optimization failures reflect agents' limited ability to create new task-specific solutions under explicit objectives. Validation failures reflect insufficient skepticism about user requests, biological assumptions, and tool capabilities. We further examine the stability of AI agents for pipeline-related tasks. According to Supplementary Figures \ref{supfig:stb} (a) and (b) for Single-Cell Omics and Supplementary Figure \ref{supfig:sp_stab} for Spatial Omics, STELLA (mem) shows relatively large variation in pipeline-based execution, while the rest of the agents have lower variation. Therefore, there is still substantial room for improving the stability, creativity, and validity-checking ability of AI agents when handling scientific tasks involving cells and tissues.

\textbf{Summary.}
Overall, our results reveal a strong-to-weak gradient in AI-agent capability that follows the agent's role relative to a known solution. Agents perform best on Data Analysis tasks, where they execute established workflows with known procedures. They perform moderately well on Model Selection tasks, but their choices are conservative and converge toward widely adopted methods such as Leiden, Harmony, scVI, Wilcoxon testing, and Moran's I. They struggle substantially on Optimization tasks, especially when required to create new predictive strategies or search for solutions beyond off-the-shelf methods. Finally, agents remain unreliable in Validation tasks for the omic study because they frequently assume that user requests are feasible rather than critically assessing scientific validity and technical executability for biomedical data.

\subsection{Benchmarking AI Agents on Processing Patient-Level Information.}

\begin{figure}
\centering
\includegraphics[width=0.95\linewidth]{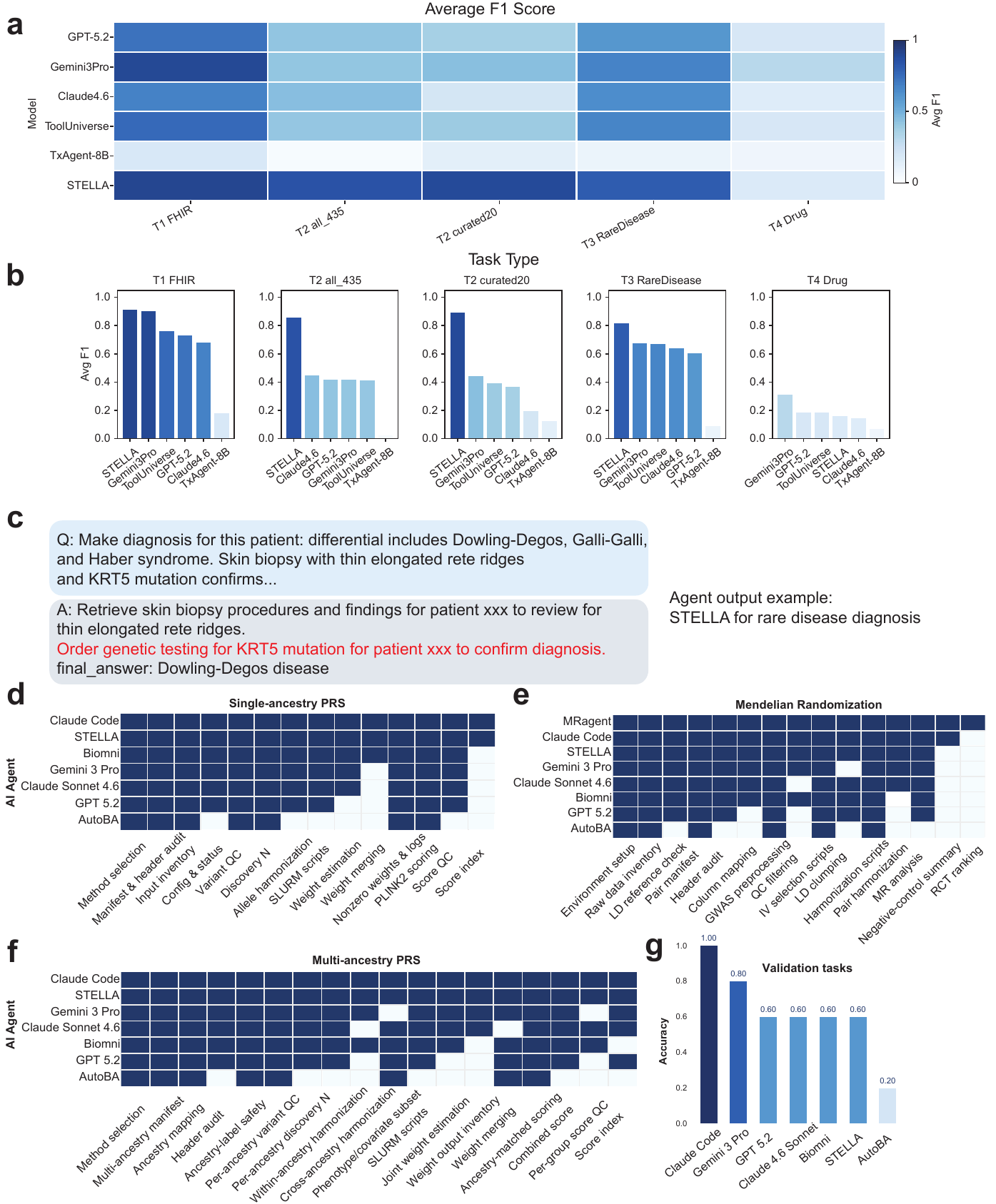}
\caption{Summary of AI agent performance on EHR-based clinical and statistical genetics tasks. (a) Average F1 score across four task types for six AI agents. (b) Per-task average F1 bar plots for T1 FHIR, T2 all\_435, T2 curated20, T3 RareDisease, and T4 Drug. (c) Case study for rare disease diagnosis with agent outcomes. For (d)-(g), panels show subtask-level outcomes for agent-generated outputs, with rows representing agents ordered by total subtasks passed and columns representing subtasks. Dark navy cells indicate pass (1), and white cells indicate fail (0). (d) Single-ancestry PRS across 25 traits within European ancestry. (e) Mendelian randomization benchmark across 3 datasets, 187 trait pairs, and 13 MR methods. (f) Multi-ancestry PRS across 22 traits and 5 ancestries. (g) Validation tasks, showing the proportion of 10 conceptual/diagnostic genetics questions answered correctly by each agent; bar shading encodes the same proportion.}
\label{fig:ehr_genetics_result}
\end{figure}

\textbf{Data Preprocessing and Analysis: agents can execute structured retrieval and standard pipelines, but performance drops as workflows become longer and more infrastructure-heavy.}
Electronic Health Records (EHRs) serve as the primary repository of \textbf{patient information} in modern healthcare, encoding clinical histories, diagnoses, medications, laboratory values, and care plans in structured, semi-structured, and unstructured formats \cite{jensen2012mining,hripcsak2013next}. Effectively navigating and reasoning over EHR data is therefore a prerequisite for any AI agent operating in a clinical environment. We evaluate six agents, including GPT-5.2, Gemini 3 Pro, Claude Sonnet 4.6, ToolUniverse, TxAgent-8B, and STELLA (mem), across EHR tasks spanning structured data retrieval, multi-step clinical workflows, rare disease diagnosis, drug management, and causal inference. Performance for T1-T4 is measured by action-level F1: each agent's response is parsed into a list of discrete predicted actions, and F1 is computed by matching individual predicted actions against the ground-truth action list via bidirectional substring matching, rather than treating the entire response as a single string (discussed in Figure \ref{fig:ehr_genetics_result} (a) and (b)).

For the Fast Healthcare Interoperability Resources (FHIR) query task (T1), agents must translate natural-language clinical questions into structured FHIR API action sequences and retrieve the correct patient-level data. This task represents a Data Analysis setting in which the correct retrieval procedure is largely known. Most agents perform strongly: Gemini 3 Pro (F1 = 0.901) and STELLA (mem) (F1 = 0.913) achieve the highest F1 scores, while GPT-5.2 (F1 = 0.731), ToolUniverse (F1 = 0.762), and Claude Sonnet 4.6 (F1 = 0.680) maintain moderate performance. By contrast, TxAgent-8B is a clear outlier (F1 = 0.179), suggesting that this smaller domain-fine-tuned model lacks the structured-query reasoning capacity required for FHIR-based retrieval. Collectively, T1 results suggest that frontier-scale general-purpose LLMs and tool-augmented agents can handle basic structured retrieval with FHIR semantics.

The stepwise clinical workflow task (T2) remains within Data Analysis because the agent is expected to reproduce a ground-truth multi-step care plan from a patient context generated by Synthea \cite{walonoski2017synthea}. However, this task is substantially harder than T1 because it requires long-horizon sequential action generation. In the 435-case evaluation set, STELLA (mem) achieves the highest F1 (0.855), while the remaining agents score substantially lower: Claude Sonnet 4.6 (F1 = 0.448) and GPT-5.2 (F1 = 0.418) are the strongest among general-purpose LLMs, followed by Gemini 3 Pro (F1 = 0.415) and ToolUniverse (F1 = 0.413), with TxAgent-8B essentially failing (F1 = 0.025). This performance gap is further confirmed in the manually curated 20-case subset, where STELLA (mem) again leads (F1 = 0.891), and all remaining agents score below 0.45. The stepwise task thus reveals a stark divide between tool-augmented agents with domain-specific capabilities and general-purpose LLMs operating without clinical grounding.

A parallel Data Analysis pattern appears in statistical genetics. Statistical genetics underpins the translation of genome-wide association studies into clinically actionable insights based on \textbf{population-level information}, from constructing polygenic risk scores (PRS) that stratify disease susceptibility to performing Mendelian randomization (MR) analyses that test putative causal relationships between exposures and outcomes. These tasks require multi-step computational workflows spanning data parsing, quality control, statistical modeling, high-performance computing orchestration, and result interpretation. We evaluate 7 agents, including Claude Sonnet-4.6, Gemini-3-Pro, GPT-5.2, ClaudeCode, Biomni, AutoBA, and STELLA (mem), across single-ancestry PRS, multi-ancestry PRS, and Mendelian randomization tasks, as well as a 10-question hallucination examination. For MR, we additionally include MRAgent \cite{MRagent}, a domain-specific agent purpose-built for MR workflows. Performance is measured by expert-designed criteria per task, each scored on a binary scale (0 or 1) (Figure \ref{fig:ehr_genetics_result}). Details of tasks and metrics are explained in the Methods section.

For the single-ancestry PRS task, agents receive GWAS summary statistics for 25 complex traits, an LD reference panel, genotype data, and phenotype/covariate files, and must produce a fully automated PRS pipeline. Claude Code and STELLA (mem) achieve perfect performance, passing all 14 subtasks and producing complete pipelines that satisfy both preprocessing and downstream scoring requirements. Biomni follows closely, passing 13/14 subtasks and failing only the final score-indexing step. Gemini 3 Pro and Claude Sonnet 4.6 each pass 12/14 subtasks; both complete the main preprocessing and scoring workflow but fail at weight merging and final score indexing. GPT-5.2 passes 11/14 subtasks, with an additional failure at weight estimation. AutoBA passes 7/14 subtasks, succeeding only on early inventory-related steps and selected scoring outputs. Overall, most agents can perform basic input auditing and variant QC, but performance diverges at infrastructure- and memory- stages, particularly weight estimation, weight merging, and reproducible score indexing.

To further test agents in more complex Data Analysis workflows, we extend the single-ancestry PRS task to a multi-ancestry PRS task. Agents are required to construct ancestry-aware PRS pipelines across 22 traits and 5 ancestry groups in UKBB. Claude Code and STELLA again achieve perfect performance, passing all 18 subtasks. Gemini 3 Pro, Claude Sonnet 4.6, and Biomni each pass 16/18 subtasks, but their failures occur at different workflow stages. Gemini 3 Pro fails phenotype/covariate subsetting and per-group score QC, suggesting that it can construct most of the ancestry-aware pipeline but struggles with evaluation-ready subgroup outputs. Claude Sonnet 4.6 fails cross-ancestry harmonization and weight merging, indicating difficulty integrating ancestry-specific resources into a unified scoring pipeline. Biomni fails weight-output inventory and final score indexing, reflecting weaker reproducibility and output-tracking infrastructure despite otherwise strong pipeline execution. GPT-5.2 passes 14/18 subtasks, with failures in cross-ancestry harmonization, joint weight estimation, weight-output inventory, and per-group score QC. AutoBA passes 8/18 subtasks, completing early method-selection and ancestry-mapping steps as well as limited downstream scoring, but failing most harmonization, SLURM, joint-weight, combined-score, QC, and indexing components. Compared with single-ancestry PRS, the multi-ancestry task exposes additional ancestry-aware failure modes: agents generally handle generic inventory and per-ancestry QC, but diverge on cross-ancestry harmonization, joint weight estimation, per-group validation, and final indexing, which are essential for evaluating PRS transferability and avoiding aggregate results that obscure ancestry-specific errors.

\textbf{Optimization (Model Selection): agents can choose familiar methods, but they converge to popular defaults rather than adaptive methodological judgment.}
Model Selection tasks require agents to choose an appropriate established method or estimator based on the data and objective. In EHR, this role appears most clearly in the causal inference task (T5), where agents are presented with an open-ended causal inference problem grounded in real EHR data: given a clinical question about whether early intravenous loop diuretic therapy reduces the risk of invasive mechanical ventilation, each agent must autonomously specify a causal estimand, construct an analytic cohort from MIMIC-IV Demo records, apply an appropriate statistical estimator, and return a machine-readable answer with verifiable artifacts. The task is evaluated across eight methodology dimensions (D1-D8; see Supplementary Figure~\ref{supfig:t5_heatmap}). Agents vary considerably in their ability to execute a reproducible analytical pipeline: frontier models with strong tool use demonstrate coherent end-to-end causal reasoning, while smaller or less capable agents fail primarily on artifact reproducibility (D6) rather than on causal conceptual understanding. Temporal validity (D3) is the one dimension all agents pass consistently, confirming that pre-/post-treatment time ordering is universally respected.

In statistical genetics, Model Selection is most visible in PRS and MR workflows. For both PRS tasks, agents show striking convergence in methodological choices, paralleling the method-selection homogeneity observed in single-cell and spatial omics. All agents select the same PRS tools, PRS-CS \cite{ge2019polygenic} and PRS-CSx \cite{Ruan2022PRSCSx}, and adopt nearly identical pipeline architectures. This indicates that agents default to the most frequently documented approaches rather than exploring alternatives that may be better suited to specific data characteristics, ancestry structure, or downstream evaluation goals. Nevertheless, agents differ substantially in implementation quality. Most agents complete basic input auditing and variant QC, while failures become more frequent in infrastructure-heavy steps, including SLURM scripting, weight estimation and merging, score QC, and final score indexing. Claude Code and STELLA show the strongest performance, benefiting from direct data access, while Biomni, Gemini 3 Pro, and Claude Sonnet 4.6 follow closely.

The MR task similarly requires agents to construct a complete summary-statistics causal-inference pipeline. Although agents are only required to successfully implement one MR method among 13 tested methods, MRAgent \cite{MRagent}, the domain-specific reference agent, is the only agent that successfully implements all 13 MR methods and achieves perfect performance by passing all subtasks. Among general-purpose agents, Claude Code performs best, passing 14/15 subtasks and failing only the final RCT-ranking component. STELLA passes 13/15 subtasks, with failures limited to the two final evidence-synthesis steps: negative-control summary and RCT ranking. Gemini 3 Pro, Claude Sonnet 4.6, and Biomni each pass 12/15 subtasks, but with different failure modes: Gemini 3 Pro fails LD clumping in addition to the final two evidence-synthesis steps; Claude Sonnet 4.6 fails QC filtering and the final two steps; and Biomni fails pair harmonization and the final two steps. GPT-5.2 passes 10/15 subtasks, with failures in column mapping, QC filtering, pair harmonization, negative-control summary, and RCT ranking. AutoBA passes 6/15 subtasks and is largely limited to early setup, inventory, and selected preprocessing components. These results indicate that most agents can initiate an MR workflow, but reliable LD-aware instrument processing, exposure-outcome harmonization, and final causal-evidence synthesis remain major bottlenecks.

Taken together, the EHR and genetics results suggest that agents can select familiar methods when the methodological space is well documented, but they often do so conservatively. Their choices converge toward popular tools or standard estimators, and performance depends heavily on whether the agent can correctly configure, execute, and document the selected method. Domain-specific agents such as MRAgent demonstrate that curated methodological knowledge and specialized tool access can substantially improve performance, especially when the task requires advanced causal-inference expertise rather than general coding ability alone.

\textbf{Optimization (Model Design): open-ended clinical and statistical decision-making remains difficult, especially when tasks require coverage, trade-off reasoning, or evidence synthesis.}
Optimization tasks require agents to construct or search for solutions that satisfy an explicit objective rather than merely reproduce a known workflow or select a familiar method. In the EHR benchmark, drug management (T4) is the clearest example. Agents must reason over concurrent medications, contraindications, drug-drug interactions, and dose adjustment. This task proves to be the most difficult across all agents and all EHR task types, with all agents achieving lower F1 scores than on other tasks. Gemini 3 Pro ranks first (F1 = 0.311), followed by GPT-5.2 (F1 = 0.187) and ToolUniverse (F1 = 0.186), while STELLA (mem) (F1 = 0.162), Claude Sonnet 4.6 (F1 = 0.146), and TxAgent-8B (F1 = 0.067) score lowest. Strikingly, STELLA (mem), which dominates T1-T3, performs comparably to the weakest general-purpose agents on T4. This indicates that even a specialized clinical agent can struggle when the task demands nuanced pharmacological reasoning, broad medication coverage, and the ability to balance competing clinical constraints.

Rare disease diagnosis (T3) occupies an intermediate position between Data Analysis and Optimization. It requires agents to integrate sparse and atypical clinical features, arrive at the correct diagnosis, and propose a rational diagnostic plan. Unlike T2, the task is less procedurally constrained and more knowledge-intensive. Agent F1 scores are substantially higher across the board: STELLA (mem) leads (F1 = 0.816), followed by Gemini 3 Pro (F1 = 0.675), ToolUniverse (F1 = 0.669), Claude Sonnet 4.6 (F1 = 0.641), and GPT-5.2 (F1 = 0.604), with TxAgent-8B remaining the weakest agent by a wide margin (F1 = 0.088). The competitive performance of ToolUniverse, despite its weaker T2 F1, suggests that when a task is better served by targeted knowledge retrieval than by sequential action planning, tool-calling frameworks can match larger general-purpose models. However, most agents cannot produce exactly matched rare disease names, and the STELLA (mem) case study in Figure \ref{fig:ehr_genetics_result} (c) shows that agents may make decisions based on impossible actions, which challenges their utility for rare disease diagnosis.

In genetics, Optimization appears in infrastructure-heavy and evidence-synthesis stages of PRS and MR workflows. Agents often complete early data auditing and standard preprocessing but fail when required to produce reproducible, evaluation-ready outputs across traits, ancestries, or causal-evidence layers. In multi-ancestry PRS, failures in cross-ancestry harmonization, joint weight estimation, per-group score QC, and final score indexing show that agents struggle when the objective is not simply to run a pipeline, but to construct a robust scoring framework that supports ancestry-specific evaluation and transferability assessment. In MR, general-purpose agents frequently fail at the final evidence-synthesis steps, including negative-control summary and RCT ranking. These components require agents to integrate causal estimates with external evidence, assess robustness, and produce a coherent causal interpretation. The advantage of MRAgent \cite{MRagent} suggests that optimization-like scientific reasoning benefits from specialized domain knowledge and curated tool support.

Overall, Optimization remains a major bottleneck across EHR and genetics. Agents can retrieve data, reproduce standard workflows, and select common methods, but they struggle when tasks require broad clinical coverage, trade-off reasoning, ancestry-aware integration, reproducible evidence synthesis, or exact diagnostic decision-making.

\textbf{Validation: agents remain unreliable at rejecting invalid premises, infeasible actions, and scientifically unsound conclusions.}
Validation tasks require agents to judge whether a task or conclusion is feasible and scientifically sound. This capability is critical in both clinical and genetic settings, where invalid actions, false causal assumptions, or unsupported diagnoses can directly compromise safety and interpretability.

In EHR, rare disease diagnosis provides an important validation challenge. Although STELLA (mem) achieves the highest F1 on T3, the case study in Figure \ref{fig:ehr_genetics_result} (c) shows that agents may make diagnostic decisions based on impossible actions. This indicates that high action-level overlap with a ground-truth diagnostic plan does not guarantee that the reasoning path is clinically executable. Similarly, drug management (T4) exposes validation failures because agents systematically underestimate the breadth of concurrent medications required for complex inpatients. This coverage shortfall indicates that agents may return incomplete medication-management decisions without adequately recognizing that the available action set or reasoning scope is insufficient.

The EHR causal inference task (T5) also contains a validation component. Agents must not only choose an estimator and construct a cohort, but also respect temporal ordering and uncertainty requirements. Temporal validity (D3) is the one dimension all agents pass consistently, confirming that agents can recognize the need for pre-/post-treatment time ordering in this setting. However, smaller or less capable agents fail primarily on artifact reproducibility (D6), showing that causal validity also depends on whether the claimed analysis can be verified through reproducible outputs. Therefore, even when agents appear to understand causal concepts, their results may remain unreliable if the supporting artifacts are incomplete or non-reproducible.

In statistical genetics, we directly assess factual reliability using 10 validation questions. The first five questions (Q1-Q5) are well-posed tasks, while the last five questions (Q6-Q10) embed subtly false methodological premises and should be refused with substantive justification. Claude Code is the only agent that scores 10/10, correctly executing all five valid tasks and correctly rejecting all five false-premise tasks with appropriate scientific justification. Gemini 3 Pro ranks second with 8/10, losing points only on Q3 and Q4 due to incorrect output, while correctly rejecting all fake tasks, indicating strong reasoning performance. GPT-5.2, Claude Sonnet 4.6, Biomni, and STELLA each reach 6/10 but with distinct error profiles: Biomni fails Q3-Q5 but correctly rejects 4 fake tasks; GPT-5.2 and Claude fail to reject 2 fake tasks; and STELLA fails to reject 3 fake tasks. This result indicates that Biomni is stronger at statistical reasoning, probably because of the domain knowledge used to train the model, while LLMs and STELLA are better at coding. AutoBA scores only 2/10, and notably, its two correct answers (Q6 and Q7) are both refusals of fake questions, while every valid question is answered with logic bugs.

These results reveal two co-existing validation failure axes: (i) silent execution bugs on valid tasks and (ii) inability to identify counterfactual or false methodological requirements, such as attempting causal inference based only on association results. This indicates that current models lack the critical scrutiny needed to serve as reliable scientific collaborators, and users must remain cautious when interpreting results generated by AI agents.

\textbf{Summary.}
Taken together, the EHR and statistical genetics benchmarks reveal a strong-to-weak gradient in AI-agent capability that mirrors the agent's role relative to a known solution. Agents perform best on Data Analysis tasks, including FHIR structured retrieval, stepwise workflows with appropriate tool augmentation, and standard PRS/MR pipeline execution. Their performance becomes more variable in Model Selection tasks, where they can select familiar methods but tend to converge toward popular defaults such as PRS-CS \cite{ge2019polygenic}, PRS-CSx \cite{Ruan2022PRSCSx}, and standard MR estimators rather than demonstrating adaptive methodological judgment. Agents struggle more substantially on Optimization tasks, including drug management, rare disease diagnostic planning, multi-ancestry PRS integration, and MR evidence synthesis, where success requires coverage, trade-off reasoning, and task-specific solution construction. Finally, agents remain unreliable in Validation tasks, where they must reject impossible clinical actions, false methodological premises, or scientifically unsupported conclusions.

This organization also clarifies why performance differs across domains and task types. In EHR, STELLA (mem) achieves the highest average F1 across all evaluated cases (F1 = 0.844), driven largely by its strong performance on T2. Among general-purpose agents, Claude Sonnet 4.6 (F1 = 0.449) and Gemini 3 Pro (F1 = 0.444) are the strongest performers, while TxAgent-8B (F1 = 0.038) consistently underperforms all other agents. The large F1 gap between STELLA (mem) and the remaining agents on T2 reflects the critical importance of tool augmentation for procedurally structured clinical tasks, whereas T3 results demonstrate that general-purpose agents relying on parametric knowledge can be competitive in knowledge-intensive diagnostic challenges. The uniformly low F1 on T4 underscores a fundamental limitation shared by all evaluated agents in pharmacological reasoning with high difficulty.

In statistical genetics, AI agents can reliably handle analysis-driven components such as code generation, data parsing, variant QC, and standard statistical workflows. However, performance degrades for tasks involving infrastructure, including SLURM execution, environment setup, reproducible indexing, and advanced statistical reasoning. The clear advantage of domain-specific agents such as MRAgent \cite{MRagent} points to a path forward: equipping AI agents with curated domain knowledge and specialized tool access can help close the gap between general coding proficiency and genuine scientific competence. More broadly, both EHR and genetics results suggest that future agents must improve not only execution accuracy, but also method adaptivity, optimization under domain constraints, and critical validation before they can be considered clinically or scientifically reliable.

\subsection{Benchmarking AI Agents on Cross-Domain Tasks.}

Our evaluation also takes into account the agent’s performance on cross-domain tasks, thereby assessing its performance on complex and multimodal tasks. We designed four cross-domain tasks, including Expression Quantitative Trait Loci (eQTL) computation \cite{gtex2020gtex}, drug target identification (TargetID) \cite{sui2026medea}, and synthetic lethality prediction (SL) \cite{sui2026medea}. The first task considers datasets from omics and genetics, while the rest of the tasks consider omics, drug discovery, and EHR modeling.

\textbf{Genetics X Omics: eQTL computation} Here we analyze our eQTL computation results from the GTEx v8 cohort, which estimates the relationship between Single-nucleotide polymorphisms (SNPs) and gene expression profiles based on Whole Genome Sequencing (WGS) and paired RNA-seq from the same individual for testing. We test the AI Agent's ability in producing codes to read the RNA-seq and WGS profiles to compute eQTL, and the observed data used for benchmarking are eQTL databases downloaded from the official GTEx v8 website. Supplementary Figure \ref{supfig:crossdomain} (a) shows the performances across four AI agents based on three metrics by considering the tested genes jointly, F1 score (F1), Jaccard Similarity (Jaccard), and PCCs between beta from two sources (PCC\_beta). Among the AI agents with reported performances, Claude Code identifies more eQTLs overlapped with observed databases compared with other AI Agents, while GPT-5.2 can produce highest PCC\_beta versus other candidates. However, all of the AI agents produce results with high standard deviation, and thus their solutions are not fair to all selected genes. Moreover, AI Agents that are not reported cannot generate codes that run successfully, highlighting the limitation of current AI Agents in reading and processing multimodal datasets. Regarding the preferences, Supplementary Figures \ref{supfig:crossdomain} (b) and (c) show the common preferences of different AI agents in method selection, as nearly all of the AI Agents choose to use Ordinary Least Squares (OLS), and most of them do not consider covariates during the computation process. Standard eQTL calculations require the inclusion of covariates, such as PCs, sex, and age, for adjustment. However, current AI agents do not support this capability.

\textbf{EHR X Drug Discovery x Omics: TargetID and SL.} Here we discuss AI Agents' capacities in addressing cross-domain tasks more relevant to biomedicine-data-driven discoveries, including TargetID and SL, adapted from Medea's released testing datasets. We randomly sample 20 tasks from their released task categories, and examine general LLMs and task-relevant AI Agents (Biomni, Medea, and ToolUniverse; TxAgent is not considered in this cross-domain task as it does not have the ability to analyze multi-omic data) in this section. According to Supplementary Figure \ref{supfig:crossdomain} (d), Medea presents the leading performance in predicting the SL for the specific A549 cell line, while ToolUniverse achieves the best performance in identifying the specific targeted gene for treatment design. At the same time, we find that general-purpose LLMs do not outperform domain-expert agents on such cross-domain tasks; therefore, an agent’s ability to read and parse datasets is a key capability for cross-domain research. However, these agents still cannot achieve particularly high scores, so improving their ability to solve cross-domain problems is also an emerging topic.

\subsection{Error Analyses and General Suggestions Summarized from Different Scientific Fields.}
During the evaluation process, we conducted a comprehensive analysis of the errors made by AI agents across domains. By synthesizing observations from Computational Drug Discovery, Single-Cell Omics, Spatial Omics, EHR, and Statistical Genetics, we identified recurring failure families that explain why agent performance decreases as tasks move from established workflow execution to method selection, optimization, and validation. Overall, agent errors are not merely coding failures. Instead, they often arise when agents must maintain workflow state, choose the correct scientific representation, verify tool assumptions, adapt methods to data-specific constraints, or reject invalid scientific premises. This error analysis, therefore, provides practical guidance for better leveraging AI agents in scientific research.

\textbf{Error Analysis: Computational Drug Discovery.} Across categories, the dominant failures are not simply failures to write code. Agents often fail when they have to choose the correct scientific representation, verify assumptions, or ground an answer in benchmark-standard references. This appears as molecular-standardization and charge-state errors in cheminformatics, brittle multi-constraint search in molecular design, incomplete safety-rule coverage and poor liability localization in safety tasks, visual and identifier-reconciliation failures in data analysis, and over-interpretation of invalid premises in the Chemical Claim Validation category. As shown in Figure~\ref{fig:dd_results} (e), an agent computes an impossible IC50 spread yet still returns a valid verdict. A second recurring pattern is premature convergence on familiar methods: agents often select plausible but generic workflows, such as standard beam search, BRICS-style \cite{degen2008brics} optimization, or incomplete local SMARTS libraries, even when the task requires a more specific reference, constraint, or validation step. Thus, the strongest agents are not simply those with the best language model, but those that sustain an executable workflow while repeatedly checking whether the data, tools, molecular representation, and scientific conclusion are valid.

\textbf{Error Analysis: Single-cell Omics.} We examine the errors produced by agents in the single-cell omic tasks and summarize them into five different categories. Supplementary Figure \ref{supfig:error_pie} (a) summarizes the major failure modes. The two dominant error categories are running errors and wrong function selection, each accounting for 33.9\% of all errors, indicating that current agents often struggle both with executable implementation and with choosing biologically or computationally appropriate analysis functions. Running errors typically arise from package incompatibilities, incorrect data structures, missing arguments, or failures in handling AnnData objects, while wrong function errors suggest that agents may misunderstand the expected workflow and select inappropriate methods. The misalignment represents 15.2\% of errors, reflecting the cases where the generated solution does not fully match the user’s requested task or evaluation criteria. Non-existent code errors account for 10.7\%, showing that agents sometimes generate unavailable functions, parameters, or APIs. Finally, misunderstanding the question contributes 6.25\%, suggesting that although agents can often parse the general task, they still occasionally fail to capture key biological constraints, dataset assumptions, or required outputs. Overall, AI agents for single-cell omics require stronger tool grounding, workflow awareness, API reliability, and task alignment to make robust and reproducible outcomes.

\textbf{Error Analysis: Spatial Omics.} We further examined the errors produced by agents in the spatial omics tasks and grouped them into four major categories. As summarized in Supplementary Figure \ref{supfig:error_pie} (b), the most frequent failure mode is non-existent code, accounting for 48.6\% of all errors. This suggests that spatial omics tasks are particularly vulnerable to API hallucination, where agents generate plausible-looking but unavailable functions, parameters, or package calls. Running errors and wrong function selection each account for 22.9\% of errors. Running errors typically arise from incorrect data structures, input type mismatches, or improper handling of spatial objects and neighborhood graphs. Wrong function errors mainly reflect cases where agents generate a spatial neighborhood graph using an executable function, but fail to account for the required symmetry of the final graph. In these cases, the code may run successfully, but the resulting graph does not satisfy the expected structure for downstream spatial analyses. Finally, out-of-memory errors represent 5.7\% of failures, indicating that some agents generate workflows that are computationally inefficient or poorly adapted to the scale of spatial transcriptomics data. Overall, the spatial error profile differs from the single-cell setting by being dominated by non-existent code rather than general runtime failures or inappropriate function choices. This pattern highlights a key limitation of current AI agents in spatial omics: they often lack reliable grounding in specialized and rapidly evolving spatial analysis APIs. Improving performance in this domain will therefore require stronger tool awareness, better validation of package functions, and more robust handling of spatial data structures and computational constraints.

\textbf{Error Analysis: EHR.}
On T2, the dominant failure mode for GPT-5.2, Gemini 3 Pro, Claude Sonnet 4.6, and ToolUniverse is over-generation rather than early truncation: predicted action sequences are often longer than ground truth, especially on hard cases, which increases false positives and limits precision, keeping F1 in a narrow mid-range despite moderate recall. TxAgent-8B shows the opposite pattern, with severe under-production and near-zero T2 F1, while STELLA avoids both extremes by combining very high recall with substantially better precision. A similar precision bottleneck appears in T1: Claude Sonnet 4.6 reaches near-perfect recall but lower precision because it appends extra clinically plausible yet non-required FHIR steps.

At the action level, \texttt{verify\_medication\_dose} in T2 is more frequently missed than upstream branching actions, suggesting weak hierarchical completion in non-STELLA agents, where the selected inference step \texttt{verify\_marker\_condition} is usually recalled because it is naturally cued as the entry step. T4 remains the hardest setting for all models, driven by two coupled issues: low recall of full medication sets and precision noise from context-inappropriate drug suggestions. In addition, STELLA shows output-format sensitivity on T4 (mixed list styles), which can further reduce string-matching scores and motivate reruns with tighter output-format constraints.

\textbf{Error Analysis: Genetics.} We further examined the errors produced by agents in the statistical genetics tasks and grouped them into four major categories. First, PRS tasks are dominated by pipeline assembly and infrastructure errors. Although agents often select reasonable PRS methods and perform basic input validation, many fail to maintain a complete executable workflow across stages. Several PRS failures arise from cross-stage contract drift, where variables, file headers, status schemas, or output formats produced by one stage are inconsistent with those expected by later scripts. Some agents also hallucinate plausible-looking tool or LD-reference paths instead of inspecting the filesystem, causing downstream stages to fail. This error is particularly severe in LLMs that don't have access to the data. This indicates that agents still struggle to treat multi-step genetics workflows as persistent systems rather than isolated coding tasks. Second, MR errors are mainly driven by API and package-grounding failures. Agents sometimes call functions with the wrong input object, use functions from the wrong R package, or rely on unavailable wrappers, particularly for less common MR methods. Third, advanced-method and evidence-synthesis steps remain difficult: agents perform better on environment setup, data inventory, instrument selection, and standard MR analysis, but fail more often in LD clumping, pair harmonization, advanced sensitivity analyses, negative-control summarization, and RCT-based ranking. Overall, the genetics error profile suggests that the main bottlenecks are multi-stage HPC orchestration, package-level API grounding, state passing across scripts, and reliable execution of advanced statistical-genetics workflows. At last, all agents showed a pervasive convergence error, in which they tend to select the most frequently documented approach instead of adapting method choice to the data.

We also analyzed the genetics validation benchmark. The dominant error is false acceptance, where agents produce a plausible analysis plan despite unsupported assumptions, such as identifying causal SNPs from summary statistics alone. Other failures include weak causal reasoning, where agents reject a flawed question for the wrong reason; compositional-prompt failure, where agents understand individual parts of a question but miss the contradiction created by combining them. In addition, we observe a data-grounding hallucination pattern, in which agents answer based on general prior knowledge rather than inspecting the provided data. For example, in Q4, LLMs such as GPT-5.2 sometimes fail to correctly identify the data type of the BMI phenotype, but assume it is continuous and proceed with an unsupported answer.

\textbf{Summary of Error Analysis.}
These findings suggest that the strongest agents are not simply those built on the most capable language models. Instead, strong scientific agents must sustain executable workflows while repeatedly checking whether the data, tools, representations, intermediate outputs, and final conclusions remain valid. Better agent design should therefore emphasize stronger tool grounding, explicit API verification, persistent state tracking across scripts, stricter output-format control, method-assumption checking, and built-in refusal mechanisms for invalid or unsupported tasks. For scientific research, agents should be used most confidently for well-specified Data Analysis workflows with verifiable outputs, more cautiously for Model Selection and Optimization, and with mandatory human oversight for Validation-sensitive tasks involving clinical safety, causal claims, molecular safety, or biological feasibility.

\begin{figure}
    \centering
    \includegraphics[width=1\linewidth]{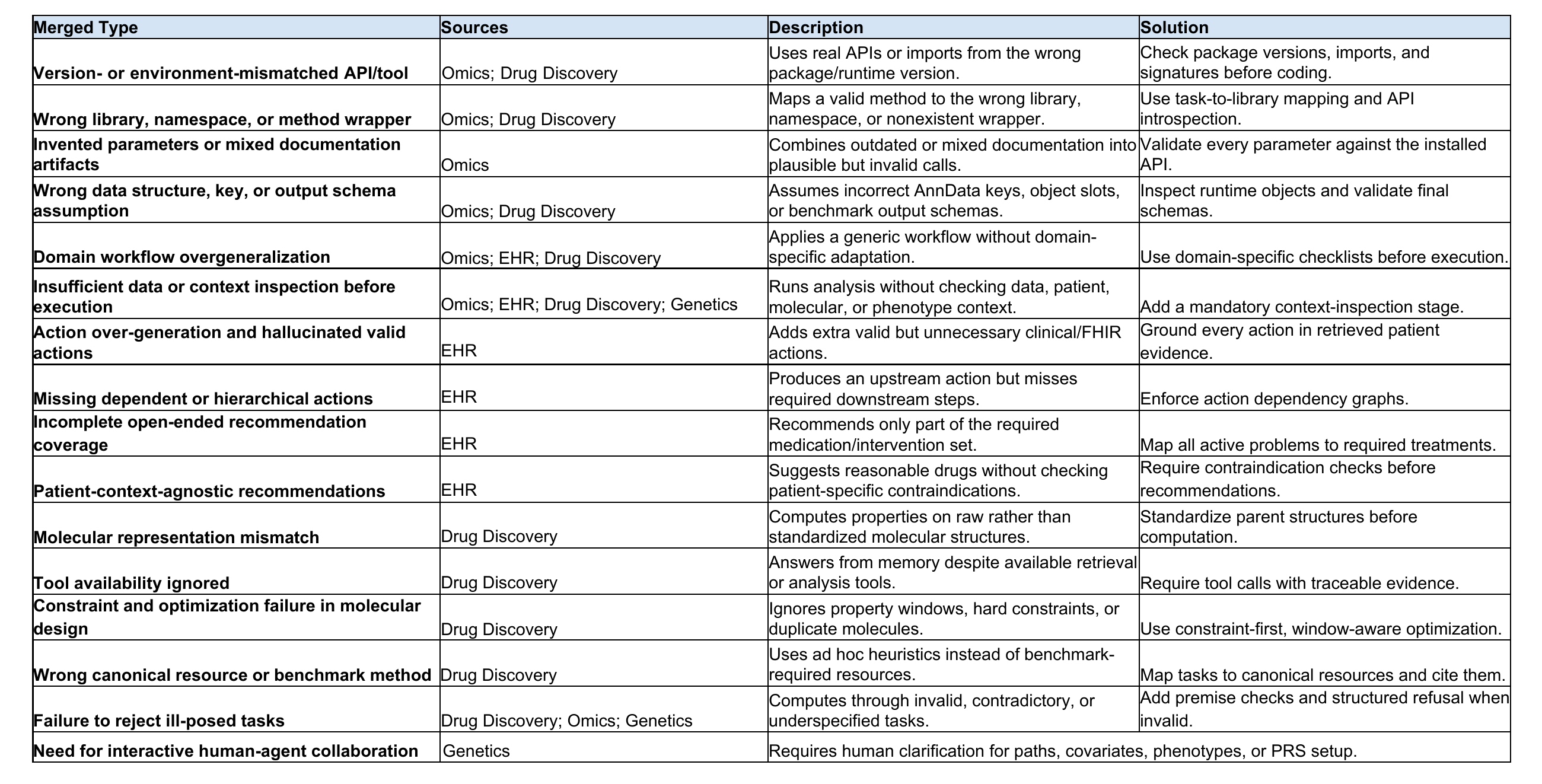}
    \caption{Summary of our benchmarking studies across different fields: Error types, sources, error descriptions, and solutions.}
    \label{fig:suggestions_and_errors}
\end{figure}

\textbf{General Suggestions.} Although our evaluation has revealed many issues with the agents, they are not entirely insurmountable. In Figure \ref{fig:suggestions_and_errors}, we outline the common problems agents encounter in scientific discovery tasks and propose feasible solutions to help users prioritize and utilize these agents more effectively. 

Across the agent benchmark, errors cluster into a set of failure modes that share a common root: the agent acts on plausible-sounding assumptions rather than verifying against the actual runtime, data, or task context before executing. This shows up most broadly in technical-execution errors, including using version- or environment-mismatched APIs, mapping methods to the wrong library or a nonexistent wrapper, inventing parameters from mixed documentation, and assuming incorrect data structures or output schemas. All of which appear across the Omics and Drug Discovery domains and point to the same remedy: check versions, imports, signatures, parameters, and runtime objects against the installed API before coding. A second cross-cutting theme is insufficient inspection and overgeneralization, where the agent applies generic workflows without domain adaptation or runs analysis without first examining the data, patient, molecular, or phenotype context; these span Omics, EHR, Drug Discovery, and Genetics, and are best addressed by mandatory context-inspection stages and domain-specific checklists. EHR errors center on clinical action management—over-generating unnecessary FHIR actions, missing required downstream or dependent steps, recommending only part of an intervention set, and ignoring patient-specific contraindications—calling for action dependency graphs, problem-to-treatment mapping, and enforced contraindication checks grounded in retrieved patient evidence. Drug Discovery errors are more chemistry- and resource-specific, involving molecular representation mismatches, constraint and optimization failures in molecular design, reliance on memory over available tools, and use of ad hoc heuristics instead of canonical benchmark resources; solutions here emphasize standardizing structures before computation, constraint-first optimization, traceable tool calls, and citing canonical resources. Finally, two distinct cross-domain failures round out the picture: failure to reject ill-posed or contradictory tasks (Drug Discovery, Omics, Genetics), which warrants premise checks and structured refusal, and the need for an interactive Co-Scientist in Genetics, where the agent should request human clarification for paths, covariates, phenotypes, or PRS setup rather than proceeding on guesses. Overall, our suggestions present as actionable fixes for the existing errors and can improve agents' performances accordingly.

We also summarize the top agents and features they present in Table \ref{tab:agent_top3_summary}. The results show a concrete strength of generalist agents, such as Claude Code, ToolUniverse, and STELLA, each of which appears in the top 3 multiple times. Claude Code wins on coding robustness, STELLA on clinical workflows, and ToolUniverse on tool-augmented tasks. Moreover, Gemini 3 Pro and GPT 5.2 also show strong performances powered by general LLM reasoning. Therefore, general-purpose agents with strong code execution and tool augmentation compete with or beat specialist agents across nearly every domain. However, under specific cases, domain-specific agents do win in their home territory. For example, MRAgent, Medea, and CACTUS also show that curated expertise still matters for sub-tasks. Therefore, rather than choosing between specialist vs. general, we suggest that the real need is for agents with better grounding and verification, regardless of domain. For example, checking APIs, inspecting data before acting, maintaining state across pipeline steps, and refusing ill-posed tasks. Those capabilities benefit every domain equally, and building them into general agents may close the gap faster than accumulating domain-specific tool libraries.

\begin{table}[H]
\centering
\resizebox{1.0\linewidth}{!}{
\begin{tabular}{ccc}
\toprule
\textbf{Agents} & \textbf{Number of Top-3 Appearances} & \textbf{Interpretation} \\
\midrule
Claude Code & 3 & Strong generalist agents, especially validity/coding-oriented tasks \\
ToolUniverse & 3 & Strong agents for biomedical domains \\
Gemini 3 Pro & 3 & Strong general LLMs \\
STELLA & 3 & Strong omics-oriented agent \\
Claude Sonnet 4.6 & 2 & Strong in perturbation/genetics \\
Biomni & 1 & Strong for eQTL estimation \\
MRAgent & 1 & Specialized MR tasks in genetics \\
GPT-5.2-Chat & 1 & Interactive chat mode which can help genetics \\
Medea & 1 & Best for SL and TargetID tasks \\
CACTUS & 1 & Best for claim validation tasks \\
\bottomrule
\end{tabular}
}
\caption{Summary of AI agents with the number of top-3 appearances or domain-specific strength as well as interpretation.}
\label{tab:agent_top3_summary}
\end{table}

\section{Discussion}
The rapid growth in both the number and capability of AI agents has created new challenges for evaluation. From the perspective of scientific research, understanding the strengths and limitations of these agents requires more than comparing different agent architectures; it also requires tasks that reflect the demands, constraints, and scenarios of real scientific problems. To enable meaningful comparisons and generate actionable insights, we introduce \method{}, a benchmark that combines step-wise verifiable, challenging scientific tasks with a broad evaluation of frontier AI agents across major scientific domains. \method{} fills the gap left by current agent benchmarks—which fail to account for different design philosophies such as self-evolving and human-AI collaboration (echoing Figure \ref{fig:benchmark_overview} (a)), and science benchmarks—which cannot address requirements of practical and challenging tasks in scientific domains (echoing Figure \ref{fig:benchmark_overview} (b)). At the same time, it provides special tasks for evaluating the limits of agent capabilities and cross-domain integration, marking a new phase in agent capacity assessment for scientific research.

Across all five domains, the central conclusion of \method{} is that current AI agents are useful but uneven scientific collaborators: they perform best when the task is a well-specified Data Analysis workflow with known procedures, such as cheminformatics preprocessing, single-cell/spatial omics pipelines, FHIR retrieval, and standard genetics analyses, but their reliability drops as tasks require longer-horizon planning, adaptive method selection, optimization, discovery, or validity checking. In drug discovery, agents can handle local molecule-processing and data-analysis tasks but struggle with multi-objective molecular optimization, mechanistic safety reasoning, and unsupported chemical claims. In single-cell and spatial omics, agents can reproduce standard workflows, yet often converge to popular defaults such as Leiden, Harmony, scVI, Wilcoxon tests, and Moran’s I, while failing when tasks require new predictive strategies, spatial-specific reasoning, or package/API awareness. In EHR modeling, tool-augmented agents are strong on structured retrieval and procedural workflows, but performance declines for rare disease diagnosis, drug management, long clinical contexts, and pharmacological reasoning. In genetics, agents can generate code, parse data, perform QC, and run standard PRS/MR pipelines, but struggle with infrastructure-heavy execution, advanced statistical reasoning, multi-ancestry integration, and evidence synthesis. Overall, we show that exploration, conservative method choice, poor state tracking, hallucinated or mismatched tools, insufficient domain validation, and a tendency to execute user requests even when the scientific premise is invalid. 

Correspondingly, we have also summarized the error reasons and proposed solutions for these issues, and thus showcase the roadmap for building the next-generation AI scientists. Future agents will need to conduct more rigorous assessments before deployment, possess a deeper understanding of the tools or skills required to solve them, and match them more accurately. They must also be able to perceive available resources and make optimal decisions within the constraints of those resources. Additionally, agents should always provide an approach for human interaction to handle special circumstances. Our conclusion thus provides the most comprehensive and timely response to the three core questions raised earlier.

\method{} follows a modular design and offers high extensibility. The code execution and evaluation components are independent of each other, allowing for simultaneous integration of agent extensions and the addition of new tasks. This enables the system to interact and coexist with the scientific community, with the goal of eventually establishing it as the ``Leetcode" of the scientific research field, as a general testing ground for agent design.

We also acknowledge several limitations. First, the tasks included in this benchmark were primarily selected by domain experts, which may have introduced unintentional but difficult-to-avoid biases. Future work should incorporate a broader range of scientific domains, such as physics and materials science, to develop more general principles for evaluating agents in scientific discovery. Second, the categories of agents evaluated in this work are considered based on existing frameworks, but the field is rapidly evolving. The benchmark should therefore be regularly updated to include emerging agent paradigms, such as multimodal LLM-based agents that can interpret and reason over scientific outputs across different modalities. These directions warrant further exploration.


\section{Methods}
\subsection{Framework} 

\textbf{Overview.} \method{} has two components, a dataset and a benchmark framework. Our dataset contains three types of questions from five different domains, and it is easy to extend. Our benchmark framework also has two components, a runtime framework and an evaluation framework. The runtime framework is designed to coordinate the running environments of different AI agents and generate the results, and the evaluation framework is designed to run evaluations based on the same environment to ensure a fair comparison. We also provide a front-end website that allows users to interact with our benchmark framework and upload their AI agents' outcomes to get evaluation results and participate in this project.

\textbf{High-level methodologies of benchmark design.} We need to design a new benchmark specifically to evaluate AI agents’ ability in solving challenging and practical problems. At the same time, we also intend to ensure that the tasks are designed in accordance with the principles of scientific research in the field, possess intrinsic connections, and can directly measure the AI agents’ ability to solve key problems. A good AI agent benchmark is grounded in this philosophy, not just a static number of simple winning rates. It emphasizes process over outcome, requiring AI agents to execute multi-step workflows with verifiable intermediate states, much like a scientist conducting experiments rather than a model generating responses. Such benchmarks prioritize falsifiability and rigor through unit-test–like evaluation of each step, incorporate tool use and environment interaction as core components of intelligence, and take the actual steps in different fields, such as omic data preprocessing, drug design, and disease diagnosis, into consideration. The tasks should span different types and difficulty levels to capture the breadth of real-world requirements, while also accounting for efficiency, cost, and other constraints to reflect resource-rational usage. Ultimately, a well-designed benchmark functions as a controlled system of the real world, testing not only capability but also generalization, reliability, and alignment. Overall, the benchmark should measure whether an AI agent can act competently, adaptively, and responsibly in complex environments, as well as provide suggestions and future directions for the next chapter of AI agent design.

\textbf{Runtime framework.} To implement the runtime framework, we implement a system to maintain and safely run frontier LLMs as well as representative AI agents in our experiments. For LLMs, we select GPT 5.2 \cite{openaigpt522025}, Claude Sonnet 4.6 \cite{anthropicclaudesonnet2025}, and Gemini 3 Pro \cite{gemini3prosystem2025} to be considered in every domain. For AI agents, we also test frontier LLMs as their default backbone models. However, since some AI agents cannot support frontier LLMs or the cost of using most advanced LLMs is huge, we choose the LLM settings recommended by the authors for those agents. Therefore, we are comparing the performance of AI agents when operating at their maximum capacity. The information on the backbone LLMs of AI agents can be found in Supplementary File 1. Our system first collects deployment methods for AI agents and LLM APIs, and then builds the corresponding environments to execute the required tasks. This design ensures that the default configurations and tool sets of different agents do not conflict with each other. Based on the task’s requirements for output (code, new data, or other structured information), we will store the AI agent’s output and use it for evaluation. We also have a time limit on the running of AI agents (no more than 24 hours). Environment control based on Conda or Mamba \cite{condacontributorscondaAsystemlevel, QuantStackandMambaContributorsmamba} also makes it easy to extend the benchmark and add new AI agents.

\textbf{Evaluation framework.} To implement the evaluation framework, we design a system to collect the outcomes from AI agents for running evaluations with a unified Input/Output (IO) interface (e.g., we can take codes and new data as inputs for the same task, and the evaluations can handle both cases). Our input data can be in different forms, as long as they can be evaluated based on the pre-defined metrics from different tasks. Our evaluation criteria are designed by experts in the relevant fields and follow the core benchmarking studies. All of these experts hold Ph.D. (or Ph.D.-level knowledge) in their respective disciplines, thereby ensuring the validity of our evaluation. The evaluation environment is configured according to the requirements of the task. We will include the packages (the most advanced version) proposed by the AI agents to ensure that the code they generate has the necessary environment to run. The output from the evaluation framework is task-specific metric scores as well as a leaderboard to rank different AI agents for comparison. We also accept other evaluation settings for the current tasks and new tasks, making it easier to extend.

\textbf{Front-end website-based system.} To enhance the framework's usability, we drew inspiration from Leetcode's design and added support for front-end access. As shown in Figures \ref{fig:platform_share} (b) and (c), we support a front-end system that enables users to specify the domains and tasks they intend to compare, and also allows users to upload various types of data for evaluation. For new users, we also provide instructions on how to expand the system, including contributing tasks or evaluating solutions, so that our system can better leverage community efforts.

\subsection{Domain: Computational Drug Discovery} 
Drug discovery unfolds as a multi-stage research enterprise that draws on a range of specialized capabilities, and progress hinges on whether each can be exercised reliably as a project evolves. Real campaigns demand rigorous molecular standardization and tool use, executable analysis across heterogeneous assay data, search and optimization against shifting multi-property objectives, screening for chemically unsafe or misleading compounds, and the scientific judgment to recognize when a conclusion outruns its evidence. Weakness in any one of these activities can corrupt downstream decisions. We therefore benchmark AI agents on drug discovery along five complementary categories.

\textbf{Definition of tasks.} We define 78 tasks across five categories. Category 1 — Chemical Data Preprocessing tests whether an agent can carry out end-to-end cheminformatics workflows reliably when molecular inputs are deliberately messy or ambiguous, demanding disciplined structure standardization and tool use. Its 17 tasks span molecular property calculation, substructure filtering, similarity ranking, format conversion, target identification, and formal charge measurement. Category 2 — Chemical Data Analysis tests whether an agent can execute sandboxed scientific data-analysis workflows over small-molecule discovery datasets. Its 26 tasks span structure-activity relationship (SAR) and dimensionality-reduction analyses; dataframe cleaning, harmonization, multi-step joins, and pipeline repair; visual interpretation of plots; and external database or tool-mediated lookup for target, assay, and ADMET enrichment.  Category 3 — Molecule Optimization tests whether an agent can adapt to heterogeneous design objectives and implement an effective optimization procedure across diverse molecular design tasks, reflecting the varied and often competing property targets that arise in real drug-discovery campaigns. Its 10 optimization tasks span rediscovery, isomer enumeration, binary-classifier objectives, narrow-band property windows, similarity-constrained property improvement, scaffold hopping, and multi-property optimization (MPO). Category 4 — Chemical Safety Assessment tests whether an agent can act as a reliable safety screener during molecular workflows, identifying, filtering, and reasoning about chemically unsafe or misleading compounds. Its 6 tasks span toxicophore structural-alert screening, hERG blocker prediction, PAINS motif screening, promiscuous-PAINS versus dark-chemical-matter classification, metabolic soft-spot and enzyme identification, and adversarial cyanide-trap molecular modification. Category 5 — Chemical Claim Validation tests whether an agent can detect when a scientific conclusion is unsupported by the underlying evidence, name the specific failure mode, and avoid overclaiming. Its 19 validity tasks pair with three of the other categories: 5 with Chemical Data Preprocessing, 11 with Chemical Data Analysis, and 3 with Chemical Safety Assessment.

\textbf{Evaluations.}
We evaluate all tasks by executing agent-generated Python code in a controlled evaluation environment and scoring the resulting outputs with task-specific procedures. In the Chemical Data Preprocessing tasks, the code is executed as a single-shot script with a fixed timeout, and its printed output is compared against benchmark reference answers using task-appropriate checks, such as numerical agreement for molecular properties, set overlap for similarity ranking, and identifier match for retrieval tasks. In the Chemical Data Analysis tasks, the code is executed in a fresh Jupyter kernel, after which an automated inspection step extracts the required variables, dataframes, and figure outputs and compares them against predefined task-specific checks; we first check whether the code runs successfully, then evaluate correctness. In the Molecule Optimization tasks, each optimizer runs as a script under a CPU-only 30-minute limit and a hard budget of 100 oracle calls; all proposed molecules are re-scored with the benchmark oracle, and we report the success rate, defined as the fraction of tasks for which the optimizer is executable, produces chemically valid molecules, and meets the formal success criterion (similarity threshold, property window, or structural-pattern match). In the Chemical Safety Assessment tasks, evaluation is tailored to the form of the safety problem: depending on the task, we compare predicted alerts, risk labels, or proposed molecular modifications against reference answers using measures such as accuracy, F1 score, AUROC, and molecule-level safety checks, with three independent runs per agent for stability. In the Chemical Claim Validation tasks, all tasks are intentionally flawed, so we score whether the agent correctly identifies the underlying failure mode, assigns the appropriate conclusion status, and recommends a sensible follow-up action, rather than rewarding a forced numerical answer. Across all five categories, this design tests not only whether an agent produces an answer, but also whether the answer is executable, chemically valid, and correct for the task at hand, serving as a challenging task.

\textbf{Datasets.}
For the Chemical Data Preprocessing tasks, we use small-molecule inputs curated to elicit structure-standardization decisions, along with structure files spanning four difficulty tiers for the formal-charge subtasks. For the Chemical Data Analysis tasks, we use a mixture of local assay and SAR tables, multi-file integration tables, and rendered visual artifacts for image-based subtasks, together with bounded cross-references to public resources including ChEMBL \cite{mendez2019chembl}, BindingDB \cite{gilson2016bindingdb}, PubChem \cite{kim2021pubchem}, and ChEBI \cite{hastings2016chebi}, and ADMET-oriented tools where the task explicitly requires them. For the Molecule Optimization tasks, we sample lead molecules from ZINC \cite{irwin2005zinc} for the property-improvement tasks and use fixed reference compounds (Celecoxib, Aripiprazole, Valsartan, Osimertinib) for the rediscovery, similarity, and MPO tasks. Scoring uses GuacaMol oracles accessed through the Therapeutics Data Commons \cite{huang2021therapeutics} library. For the Chemical Safety Assessment tasks, we use datasets supporting toxicophore structural alerts, hERG bioactivity model training and evaluation, PAINS filtering, promiscuous-PAINS versus dark-chemical-matter classification, metabolic soft-spot identification with enzyme labels, and cyanide-trap molecular modification. For the Chemical Claim Validation tasks, we use curated tabular and structural inputs designed to instantiate each canonical failure mode, with a subset of tasks requiring bounded reconciliation against public sources such as ChEMBL, PubChem, and ChEBI. None of these datasets is used to train or fine-tune the agents.

\subsection{Domain: Single-Cell Omics} 
Studying single-cell omics, particularly single-cell RNA sequencing (scRNA-seq), is crucial because it reveals the true cellular diversity and dynamic states that are masked in bulk measurements, enabling researchers to identify rare cell populations, reconstruct developmental and disease trajectories, and pinpoint the specific cell types driving biology or disease mechanisms. Capturing gene expression at single-cell resolution allows for the precise dissection of complex systems, such as the immune system, tumor microenvironment, and brain tissue, which is essential for understanding the mechanisms of diseases like cancer and neurodegeneration \cite{skinnider2025clinical, jovic2022single}. Moreover, when integrated with other information (e.g., genetic and chemical perturbations), single-cell omics can provide a more comprehensive view of understanding gene regulation and drug responses, supporting the development of targeted therapies and advancing precision medicine, while also serving as a powerful foundation for AI-driven discovery in high-dimensional biological data. Therefore, we study AI agents' contributions in addressing challenges related to single-cell omics analysis and optimization.

\textbf{Definition of tasks.} We have 12 different tasks from three categories to benchmark AI agents' ability in the study of data analysis and algorithm design. The first category is to design a preprocessing pipeline for scRNA-seq data. Here we have seven tasks designed to examine the ability of AI agents in preprocessing the original raw count data matrix and converting the matrix to a normalized version, including READ, QC, Basic Filtering (FILTER), Doublet Detection (DOUBLET), Normalization (NORM), Highly Variable Gene Selection and Principal Component Analysis (HVG$+$PCA), and Neighborhood graph construction$+$UMAP (NEG+VIS). These tasks assess the most fundamental capabilities of AI agents in processing single-cell data, including steps such as reading data, performing quality control, and dimensionality reduction. We also have three tasks designed to evaluate AI agents' ability in selecting the optimal solution in the design of the preprocessing pipeline, which include the optimal selection of clustering method, batch effect correction method, and Differentially Expressed Genes (DEGs) detection. We require AI agents to utilize their information extraction and data analysis capabilities to select the most appropriate method for executing tasks. We also have one optimization task, which requires AI agents to design the optimal solution to integrate scRNA-seq data from different developmental stages and keep the trajectory information, as well as one optimization task which requires AI agents to design an algorithm to predict the perturbation effect based on unseen perturbation labels based on perturb-seq datasets. Our design also takes into account the difficulty of the problems; optimization-related problems tend to be more challenging, particularly the task involving perturbation prediction.

For these tasks, we provide expert solutions as references. The reference codes for pipeline design are developed based on the tutorial of Scanpy \cite{wolf2018scanpy}, and the method for perturbation prediction is scLAMBDA \cite{wang2024modeling}, which leverages LLM embeddings to improve prediction performance.

We also have 15 questions to benchmark AI agents' ability to identify whether the task can be executed (validity examination). We have ten easy questions and five hard questions, which are all relevant to the real challenges in the study of single-cell omics (e.g., Identifying major cell types and related marker genes, and Inferring large-scale chromosomal Copy Number of Variants (CNVs) across endocrine cell populations), and require AI agents to make an answer based on diving deep into the dataset. 

\textbf{Evaluations.} For the analysis-related tasks, we use pass$@$1 to evaluate. We check if the data after processing, based on the proposed solutions from AI agents, passes all checkpoints or not. For the batch effect correction optimization task, we refer to scIB \cite{luecken2022benchmarking} and compute $S_{total}$ by considering the metrics for evaluating batch effect reduction and biological information conservation. If the method failed in producing a usable algorithm, we compute PCs for evaluating batch effect correction. For the clustering optimization task, we use classical clustering-related metrics \cite{pedregosa2011scikit} to evaluate the similarity between proposed clusters and cell types, including Adjusted Rand Index (CLS\_ARI), Adjusted Mutual Information (CLS\_AMI), Normalized Mutual Information (CLS\_NMI), Homogeneity Score (CLS\_Homo), Completeness Score (CLS\_Completeness), V-measure (harmonic mean of homogeneity and completeness) score (CLS\_V-measure), and Fowlkes–Mallows Index (CLS\_FMI), and Purity score(CLS\_Purity). For the DEG detection task, we collect publicly available marker genes for cell types as observed labels and evaluate the similarity between algorithm outputs and collected labels. Our metrics \cite{pedregosa2011scikit} include Jaccard similarity (DEG\_jaccard), Overlap coefficient (DEG\_overlap\_coef), Precision rate (DEG\_precision), Recall rate (DEG\_recall), and F1 rate (DEG\_F1). For the optimization of trajectory integration, we consider the trajectory preservation score from scIB as the metric, which computes the similarity between the observed pseudotime and integrated cell embeddings. For perturbation prediction, we compare the similarity between predicted and observed gene expression profiles at the perturbation-label level (we average the cells from the same condition to compute metrics), including Pearson Correlation Coefficients for all genes (PCC\_all) and top\_20 DEGs (PCC\_DEGs), as well as Mean Squared Error (MSE). These metrics have been corrected by subtracting the mean expression level of the control cells. For the task of validity check, we report accuracy based on comparing the ground truth answers and agent-produced answers. All of the metrics, except MSE, are in the range between 0 and 1. Higher scores mean better methods. For MSE, the score range is 0 to infinity, and lower scores represent better methods.

\textbf{Datasets.} For the task of pipeline design, we select three scRNA-seq datasets from three different tissues: Pancreas \cite{baron2016single, muraro2016single, segerstolpe2016single, wang2016single, xin2016rna}, PBMC \cite{zheng2017massively}, and Heart \cite{litvivnukova2020cells}. All datasets come from humans and are annotated with cell types. Heart is an atlas-level dataset with a large scale. We collect the marker genes for cell types in these tissues from CellMarker 2.0 \cite{hu2023cellmarker}. For the task of trajectory integration, we select two datasets with developmental stages. One from the HSC cell series, the other from the B cell series \cite{inecik2025beyond, suo2022mapping,liu2024evaluating}. For the task of perturbation prediction, we select three perturb-seq datasets, including Norman \cite{norman2019exploring}, Adamson \cite{adamson2016multiplexed}, and RPE1 \cite{replogle2022mapping}. These datasets are not used for LLM training.

\subsection{Domain: Spatial Omics}
Spatial omics has become an essential modality because it preserves the spatial organization of tissue while measuring molecular profiles, allowing researchers to connect gene expression to anatomical structure, local cellular neighborhoods, and tissue-level function in ways that dissociated single-cell assays cannot \cite{Fischer2023, Kuemmerle2024, Yang2025}. This spatial context is critical for identifying biologically meaningful tissue domains, resolving neighborhood organization and enrichment patterns, and detecting spatially variable genes that mark anatomical boundaries or region-specific programs across development and disease. As spatial datasets continue to grow in scale, complexity, and clinical relevance, robust performance on these core preprocessing, analysis, and optimization tasks becomes increasingly important. Therefore, we benchmark AI agents on spatial omics tasks that reflect these fundamental analytical demands.

\textbf{Definition of tasks.} We have 11 different tasks from two categories to benchmark AI agents' ability in the study of data analysis and algorithm design. Similar to the single-cell domain, the first category is to design a preprocessing pipeline for spatial transcriptomics data. Here we have eight tasks designed to examine the ability of AI agents in preprocessing the original raw count data matrix and the spatial coordinates, including READ, QC, Basic Filtering (FILTER), Normalization (NORM), Highly Variable Gene Selection and Principal Component Analysis (HVG$+$PCA), Neighborhood graph construction$+$UMAP (NEG+VIS), Spatial Neighborhood graph construction (SNG), and Spatial Neighborhood Enrichment Analysis (NEA). These tasks assess the most fundamental capabilities of AI agents in processing spatial transcriptomics data, including steps such as reading data, performing quality control, dimensionality reduction, and spatial neighborhood setup. Similar to the single-cell domain, we also have three tasks designed to evaluate AI agents' ability in selecting the optimal solution in the design of the preprocessing pipeline, which include the optimal Batch Effect Correction (BEC) method, Spatially Aware Domain (SAD) detection, and Spatially Variable Genes (SVGs) detection. We require AI agents to utilize their information extraction and data analysis capabilities to select the most appropriate method for executing tasks. Our design also takes into account the difficulty of the problems; optimization-related problems tend to be more challenging, usually due to hallucination errors.

For these tasks, we provide expert solutions as references. The reference codes for pipeline design are developed based on the tutorial of Squidpy \cite{Palla2022squidpy}.

We also designed a 15-question benchmark, with some questions paired with proposed analysis pipelines, to evaluate whether AI agents can correctly determine if a requested task is actually executable on the given dataset. Four of the questions are valid and reflect realistic spatial transcriptomics analyses, such as identifying spatial domains or examining layer-specific marker genes. The remaining questions are intentionally flawed and are designed to test robustness. These flawed cases include, for example, questions paired with hallucinated or erroneous code, incorrect species assumptions, or requests that cannot be supported by the available data modality or study design. To answer them correctly, an AI agent must go beyond surface-level pattern matching and carefully inspect both the dataset and the proposed pipeline before deciding whether the task is valid.

\textbf{Evaluations.} For the analysis-related tasks, we use pass$@$1 to evaluate. We check if the data after processing, based on the proposed solutions from AI agents, passes all checkpoints or not. For the batch effect correction optimization task, we refer to scIB \cite{luecken2022benchmarking} and compute $S_{total}$ by considering the metrics for evaluating batch effect reduction and biological information conservation. If the method failed in producing a usable algorithm, we compute PCs for evaluating batch effect correction. For the spatially aware domain optimization task, we use classical clustering-related metrics \cite{pedregosa2011scikit} to evaluate the similarity between proposed spatial domains and cell types, including Adjusted Rand Index (SAD\_ARI), Adjusted Mutual Information (SAD\_AMI), Normalized Mutual Information (SAD\_NMI), Homogeneity Score (SAD\_Homogeneity), Completeness Score (SAD\_Completeness), V-measure (harmonic mean of homogeneity and completeness) score (SAD\_V-measure), and \\ 
Fowlkes–Mallows Index (SAD\_FMI), and Purity score(SAD\_Purity). We also calculate the Global Moran's I score for spatial domain labels to assess similarity between labels of nearby spots \\ (SAD\_Global\_Moran\_I). For the SVG detection task, 
since no ground truth is available, we evaluate the similarity between the algorithm's outputs and a curated set of silver-standard SVGs referencing the criteria defined in a benchmarking paper \cite{Chen2025_silver_svg}. Our metrics \cite{pedregosa2011scikit} include Jaccard similarity (SVG\_jaccard), Overlap coefficient (SVG\_overlap\_coef), Precision rate (SVG\_precision), Recall rate (SVG\_recall), and F1 rate (SVG\_F1). For the task of validity check, we report accuracy based on comparing the ground truth answers and agent-produced answers. All of the metrics are in the range between 0 and 1. Higher scores mean better methods.

\textbf{Datasets.} We use three spatial transcriptomics datasets from three different tissues: Human dorsolateral prefrontal cortex \cite{Louise2024human_dlpfc} with 12 samples, accessible through the R package \texttt{spatialLIBD} \cite{spatialLIBD}; Mouse embryo with 10,000 spots randomly subsetted from 4 samples at E16.5 \cite{Chen2022stereo_seq}; and Human breast with 11 samples \cite{COUTANT2023_human_bc}. All datasets are annotated with cell types. These datasets are not used for LLM training. 

\subsection{Domain: Electronic Health Records}

Electronic health records (EHRs) serve as the primary repository of longitudinal patient information in modern healthcare, encoding clinical histories, diagnoses, medications, laboratory values, imaging orders, and care plans in structured, semi-structured, and unstructured formats anchored to standardized vocabularies and interoperability standards such as the Fast Healthcare Interoperability Resources (FHIR) protocol \cite{mandel2016smart, bender2013hl7}. The ability to navigate, reason over, and act upon EHR data is therefore a prerequisite for any AI agent operating in a clinical environment \cite{rajpurkar2022ai, topol2019deep}. Unlike single-turn question-answering, EHR tasks routinely require agents to execute multi-step procedural workflows, constructing valid API queries, verifying medication orders against patient-specific conditions, and synthesizing diagnostic plans from sparse or atypical clinical evidence, where partial solutions carry direct safety implications and the cost of both over-production (hallucinating unrequired clinical orders) and under-production (omitting necessary care steps) must be jointly minimized \cite{hripcsak2013next, obermeyer2019dissecting}. Moreover, EHR tasks span a qualitative spectrum from procedurally constrained retrieval and workflow execution to knowledge-intensive diagnostic reasoning and open-ended pharmacological decision-making, making EHR an unusually demanding domain for evaluating agents across multiple capability dimensions simultaneously. We therefore study AI agents' contributions in addressing the challenges of structured clinical data retrieval, multi-step workflow execution, rare disease diagnosis, and drug management.

\textbf{Definition of tasks.} We design four EHR task types comprising 505 evaluable patient cases to benchmark AI agents' ability in clinical data processing, multi-step action planning, knowledge-intensive diagnostic reasoning, and pharmacological decision-making. We evaluate six LLMs and agents across all tasks: GPT-5.2, Gemini 3 Pro, Claude Sonnet 4.6, ToolUniverse, TxAgent-T1-LLaMA-3.1-8B \cite{gao2025txagentaiagenttherapeutic}, and STELLA.

The first task type is the FHIR clinical workflow task (T1, $n=20$), which requires agents to translate natural-language clinical questions into structured FHIR API action sequences and correctly retrieve or create the required patient-level resources. We construct 20 patient cases spanning hypokalemia management, diabetes management, acute kidney injury due to nephrotoxic exposure, sepsis bundle initiation, and a combined hypokalemia-diabetes workflow. The ground-truth action vocabulary for T1 comprises four FHIR operations: \texttt{fhir\_search\_observation}, \\ 
\texttt{fhir\_create\_medication\_request}, \\ \texttt{fhir\_create\_service\_request}, and \texttt{fhir\_search\_procedure}. The average ground-truth sequence length is 4.05 actions per case. We stratify cases by difficulty according to sequence length: easy (2-3 steps, $n=6$), medium (3-4 steps, $n=4$), and hard (5-6 steps, $n=10$).

The second task type is the stepwise clinical workflow task (T2, $n=455$), which requires agents to produce a complete sequence of clinical decisions and verification actions aligned with a ground-truth care plan for a given patient context. The ground-truth action vocabulary comprises four operations: \texttt{verify\_marker\_condition}, \texttt{branch\_decision}, \texttt{verify\_medication\_order}, and \texttt{verify\_medication\_dose}. The main evaluation set (\texttt{all\_435}, $n=435$) is stratified into three difficulty tiers: easy (approximately 4 steps, $n=150$), medium (approximately 5 steps, $n=140$), and hard (approximately 7 steps, $n=145$), reflecting progressively longer action chains and dual-pathway branching structures. 

The third task type is the rare disease diagnosis task (T3, $n=20$), which requires agents to identify the rare disease given sparse and atypical clinical features and to propose a rational diagnostic and treatment plan expressed as an action sequence. We design 20 cases each corresponding to a distinct rare disease, including sarcoidosis, acute myeloid leukemia, Brugada syndrome, progeria, TAFRO syndrome, Kimura disease, and others. The ground-truth action vocabulary follows the FHIR resource format used in T1: \texttt{fhir\_search\_observation}, \texttt{fhir\_create\_condition}, and \texttt{fhir\_create\_service\_request}. Cases are stratified as easy ($n=6$), medium ($n=7$), and hard ($n=7$) according to the clinical complexity and rarity of the target diagnosis.

The fourth task type is the drug recommendation task (T4, $n=10$), which requires agents to recommend the correct set of medications for an inpatient given the patient's documented diagnoses and key laboratory values. Unlike T1-T3, the output is an open-ended medication list rather than a structured action sequence, and agents must produce drug names with sufficient lexical fidelity for automated scoring. This task is closest to clinical decision support for pharmacological management, demanding integration of disease-specific treatment guidelines, drug-drug interaction awareness, and patient-specific contraindication reasoning.

T5 poses a target-trial-style causal question on MIMIC-IV Demo (v2.2): does early intravenous loop diuretic therapy (within 24\,h of admission) reduce the risk of invasive mechanical ventilation during hours 24-96 post-admission among adults admitted for suspected fluid overload? Agents must autonomously define a causal estimand, construct an analytic cohort from raw EHR tables, adjust for pre-treatment confounders, apply a valid causal estimator, and return a structured \texttt{answer.json} together with a \texttt{cohort.csv} and analysis script. Responses are scored on eight dimensions (D1-D8, $\text{max}=200$) using an ordinal scale $\{0,1,2\}$: estimand specification (D1, 20\,pts), cohort construction (D2, 30\,pts), temporal validity (D3, 30\,pts; hard gate), confounder coverage (D4, 20\,pts), method correctness (D5, 30\,pts), artifact reproducibility (D6, 30\,pts), result accuracy (D7, 20\,pts), and robustness/uncertainty reporting (D8, 20\,pts). D3 disqualifies submissions that use post-treatment data for baseline variables; D6 recomputes the point estimate from the submitted cohort file. Free-text fields in D1-D2 are graded by GPT-4o-mini; remaining sub-criteria are rule-based.

\textbf{Evaluations.} For T1-T3, we use action-level F1 as the evaluation metric, computed via bidirectional substring matching between the agent's predicted action list and the ground-truth action sequence. Precision is defined as the fraction of predicted actions that match at least one ground-truth action; recall is the fraction of ground-truth actions matched by at least one predicted action; F1 is their harmonic mean. These metrics are macro-averaged across cases within each task group. For T4, we replace substring matching with fuzzy string matching (Python \texttt{SequenceMatcher} ratio $\geq 0.82$) to accommodate surface-form variation in drug names (e.g., brand names, abbreviations, and generic name variants), and apply the same F1 computation. All metrics are in the range $[0, 1]$; higher scores indicate better performance.

\textbf{Datasets.} For T1 and T2, we generate patient cases using Synthea \cite{walonoski2017synthea}, an open-source synthetic patient generator that produces clinically realistic longitudinal EHR data anchored to evidence-based clinical pathways. Synthea outputs are represented in FHIR R4 format, enabling direct evaluation of FHIR API action sequences. For T3, we construct cases using the Synthea rare-disease patient profiles augmented with manually curated clinical feature sets that reflect the diagnostic criteria and typical presentations of each target rare disease. For T4, we use ten inpatient cases from the MIMIC-IV Clinical Database Demo \cite{johnson2023mimic}, a publicly available de-identified subset of the MIMIC-IV database derived from real intensive-care and general hospital admissions at Beth Israel Deaconess Medical Center. The ground-truth medication lists for T4 are derived from the actual inpatient prescriptions documented in the MIMIC-IV records. These datasets are not reported to be used for training any of the six evaluated language models.

\subsection{Domain: Genetics}

Genetics provides a central foundation for understanding disease risk, causal mechanisms, and population-specific biomedical variation. Modern statistical genetics workflows require agents to integrate heterogeneous genomic, phenotypic, and summary-statistics data; implement reproducible analysis pipelines; select appropriate statistical models; and interpret results under realistic assumptions about ancestry, linkage disequilibrium, confounding, and causal validity. These workflows are challenging for AI agents because they require both procedural correctness and domain-specific reasoning. Errors in data harmonization, allele alignment, population matching, covariate adjustment, or causal interpretation can substantially distort downstream conclusions. We therefore benchmark AI agents on four complementary genetics task categories: single-ancestry polygenic risk score prediction, Mendelian randomization, multi-ancestry polygenic risk score prediction, and conceptual validation.

\textbf{Definition of tasks.} We design genetics tasks to evaluate whether AI agents can perform core statistical genetics analyses from input data and task instructions. The first task category is single-ancestry polygenic risk score (PRS) analysis (T1, $n=14$). This task evaluates whether agents can construct a reproducible European-ancestry-based PRS score-generation pipeline end-to-end from raw GWAS summary statistics and target genotypes. In particular, agents receive a trait manifest pointing to per-trait GWAS sumstats and genotype/phenotype data from UKBB European Samples. Then they are required to select one optimal PRS computation method (i.e., PRScs, LDpred2, etc.) and deliver four staged sub-task groups: (i) method selection and input validation; (ii) variant QC and allele harmonization; (iii) PRS weight estimation; and (iv) PRS coring and score QC. The benchmark also highlights whether agents can correctly handle the classical PRS failure modes — allele flips, strand ambiguity, rsID/CHR: POS identifier substitution, discovery-vs-target N confusion, per-chromosome job indexing, and silent variant loss during harmonization or merging.

The second task category is Mendelian randomization (MR) analysis (T2, $n = 15$), which evaluates whether agents can estimate causal effects from genetic instruments using GWAS summary-statistics data. Following the study design of Hu et al. \cite{Hu2024MRBenchmark}, we construct a 187-pair MR benchmark across three datasets that probe distinct failure modes, evaluated under 13 MR methods. Dataset 1 ($n=100$) evaluates whether agent can identify spurious causal effect induced by population-stratification, using 20 representative UK Biobank exposures from the Neale Lab with 5 pigmentation outcomes (four hair-colour categories and ease of skin tanning) that exhibit strong north--south genetic gradients in European populations; Dataset 2 ($n=77$) evaluates whether agent can identify spurious causal effect induced by horizontal-pleiotropy, using 11 adult-behaviour and aging-related exposures with 7 childhood-outcome traits; Dataset 3 evaluates whether agents recover known causal effects on well established trait pairs including LDL-CAD, BMI-CAD, BMI-T2D and SBP-CAD. For each trait pair, agents are required to harmonize raw exposure and outcome summary statistics, perform LD clumping to select valid genetic instruments and apply 13 MR methods: inverse-variance weighted (IVW) with fixed and random effects \cite{Burgess2013IVW}, debiased IVW (dIVW) \cite{Ye2021dIVW}, MR-Egger \cite{Bowden2015MREgger}, MR-RAPS \cite{Zhao2020MRRAPS}, weighted median \cite{Bowden2016WeightedMedian}, weighted mode \cite{Hartwig2017WeightedMode}, MR-PRESSO \cite{Verbanck2018MRPRESSO}, MRMix \cite{Qi2019MRMix}, cML-MA \cite{Xue2021cMLMA}, MR-Robust and MR-Lasso \cite{Rees2019MRRobust}, and MR-ConMix \cite{Burgess2020MRConMix}. We exclude three methods due to the high computational burden for CAUSE and MR-APSS, and the lack of public UK10K LD-block resources required by MR-CUE. Agents are required to return structured per-pair estimates comprising the point estimate standard error, 95\% confidence interval, and significance measure.

The third task category is multi-ancestry PRS analysis (T3, $n = 18$). This task extends the single-ancestry PRS benchmark to a multi-ancestry setting covering 22 traits (GLGC lipids, PAGE anthropometric and cardiometabolic traits, BBJ blood and liver biomarkers) and up to five UK Biobank ancestry target cohorts (EUR, EAS, AFR, SAS, AMR) as used in \cite{Xu2025JointPRS}, with per-trait ancestry coverage explicitly heterogeneous (e.g., GLGC traits span all five ancestries, PAGE traits span EUR+EAS+AFR, BBJ traits span EUR+EAS). Similarly, agents receive the ancestry-aware manifest including trait, ancestry, and summary statistics. Then they are asked to select one optimal joint multi-ancestry Bayesian method (i.e., PRS-CSx, JointPRS, SDPRX, etc.), justify the choice, and then complete four sub-task groups: (i) method selection, ancestry/genotype mapping, and input validation; (ii) per-ancestry preprocessing and harmonization; (iii) joint PRS weight estimation; and (iv) ancestry-matched scoring. The benchmark tests whether agents can select joint rather than EUR-only methods, propagating ancestry-specific N and LD, refusing to collapse heterogeneous SNP sets, and producing per-ancestry PRS scores in a layout that supports downstream evaluation of cross-ancestry transferability — while avoiding multi-ancestry-specific failure modes, including silent ancestry drop, target-cohort N substitution, cross-ancestry weight misapplication, and unmapped ancestry labels.

The last task category is genetics validation, which evaluates conceptual and diagnostic reasoning rather than code execution alone. Each agent answers 10 expert-designed genetics questions covering common analytical assumptions, diagnostic checks, and interpretation challenges in statistical genetics. The first five questions are designed to be valid and answerable, and we compare each agent's responses against expert-curated ground-truth labels to report the proportion answered correctly. The remaining five questions are intentionally invalid or fabricated, and we evaluate whether the agent can recognize the unsupported premise, reject the question appropriately, and provide a plausible justification.

\textbf{Evaluations.} For the single-ancestry PRS, Mendelian randomization, and multi-ancestry PRS tasks, we evaluate agent outputs at the subtask level. Each subtask is scored as pass or fail, where pass indicates that the submitted output satisfies the predefined correctness criteria for that subtask. These criteria include successful execution, correct data processing, valid output formatting, and agreement with reference results or accepted analytical checks. For PRS tasks, evaluation focuses on whether the agent produces valid PRS outputs and correctly reports predictive performance according to the required metric. For MR tasks, evaluation checks whether agents correctly harmonize data, apply the specified MR methods, and return causal-effect estimates in the required format. Due to the large number of MR methods being benchmarked, we only require agents to successfully implement and interpret at least one MR method to pass the task. For multi-ancestry PRS tasks, evaluation additionally verifies ancestry-stratified outputs.

For the validation tasks, we compare each agent’s responses to the first five questions against expert-curated ground-truth labels and report the proportion answered correctly. For the remaining five questions, we evaluate whether the agent can recognize bias or invalid assumptions embedded in the question and appropriately reject the premise with a plausible justification. 

\textbf{Datasets.} For single-ancestry PRS and multi-ancestry PRS, we utilized genotype data and phenotype data from UK Biobank\cite{UKBB} and 1000G\cite{1000G} as the LD reference panel. Summary statistics used for single-ancestry PRS are derived from UKBB data, while summary statistics used for multi-ancestry PRS are publicly available as indicated by Xu et al.\cite{Xu2025JointPRS}. For Mendelian randomization, the 25 summary statistics of dataset 1 are downloaded from the Neale Lab UK Biobank GWAS \cite{NealeLab2018}; the 18 summary statistics of dataset 2 are downloaded from the GWAS Catalog \cite{Vogelezang2020ChildBMI}, the Social Science Genetic Association Consortium (SSGAC) \cite{KarlssonLinner2019SSGAC}, the Neale Lab UK Biobank GWAS \cite{NealeLab2018}, the IEU OpenGWAS database \cite{deelen2019meta, timmers2019genomics}, Zenodo \cite{zenin2019identification}, the University of Bristol Research Data Repository \cite{kuo2020dupuytren}, figshare \cite{atkins2021genome}, the EArly Genetics and Lifecourse Epidemiology (EAGLE) consortium \cite{middeldorp2019early}, the Childhood Intelligence Consortium (CHIC) \cite{benyamin2014childhood}, the Early Growth Genetics (EGG) consortium \cite{middeldorp2019early}, and GWAS Atlas \cite{tian2020gwas}; the 11 summary statistics of dataset 3 are downloaded from the Global Lipids Genetics Consortium (GLGC) \cite{dron2025breadth}, the Neale Lab UK Biobank GWAS \cite{NealeLab2018}, the Genetic Investigation of ANthropometric Traits (GIANT) consortium \cite{yengo2018meta}, the International Consortium of Blood Pressure (ICBP) \cite{simino2014gene}, the DIAbetes Genetics Replication And Meta-analysis (DIAGRAM) consortium \cite{xue2018genome}, and the Coronary ARtery DIsease Genome wide Replication and Meta-analysis plus the Coronary Artery Disease Genetics consortium (CARDIoGRAMplusC4D) \cite{mcpherson2016genetics}.

\subsection{Cross Domain Tasks.}

\textbf{eQTL computation.} Expression Quantitative Trait Loci (eQTLs) \cite{nica2013expression} are genomic regions—specifically genetic variants such as
SNPs—that influence the expression levels of one or more genes. By linking genetic variation to gene expression, eQTL analysis bridges the gap between genotype and phenotype, aiding in the understanding of disease mechanisms, typically identifying cis-eQTLs (nearby genes) or trans-eQTLs (distant genes). Here, we consider the general computation accuracy of eQTL based on the GTEx v8 dataset, and require AI agents to generate codes based on prompting, RNA-seq dataset from blood tissue, and WGS data as inputs. We randomly select 100 genes to improve the efficiency of evaluation, and report the mean score and standard deviation across genes.

We download ground-truth eQTLs from the website of GTEx, which is computed based on linear regression with covariate correction based on the same cohort. Our metrics \cite{pedregosa2011scikit} include F1 score (F1), Jaccard Similarity (Jaccard), and PCCs between the beta from two different sources (PCC\_beta). Higher scores mean better methods.

\textbf{drug target identification.} This task requires AI agents to analyze transcriptomic data and select the most possible targeted gene among five candidate genes of one drug. We subsample 20 questions from the Medea-released benchmark and provide the environment and datasets for all tested agents to perform the task. We utilize Accuracy to make comparisons. Higher scores mean better methods.
 
\textbf{synthetic lethality prediction.} This task requires AI agents to predict if the modification of specific genes will lead to cell lethality or not. We subsample 20 questions from the Medea-released benchmark and provide the environment and datasets for all tested agents to perform the task. We utilize Accuracy to make comparisons. Higher scores mean better methods.

\section{Code and Data Availability}
We use the authorized OpenAI API, Claude API, and Gemini API to develop our method and perform benchmark. We also use the Misha cluster from Wu Tsai Institute at Yale, the McCleary cluster from Yale Research Computing Center, Discovery Center from Northeastern University to finish experiments. Our codes can be found in the GitHub Repo (link: \url{https://github.com/HelloWorldLTY/SciAgentArena}).

All datasets used in this study are publicly available. The complete collection of tasks/solutions, datasets, associated metadata, and configurations that constitute the \method{} benchmark is hosted in Huggingface (link: \url{https://huggingface.co/datasets/iLOVE2D/SciAgentArena}).

\section{Author Contributions}

Coordination and planning: Tianyu Liu (lead), Yuanqi Du, Wengong Jin, Pan Lu, Zhuoran Yang, Kaize Ding;
Framework design and development: Tianyu Liu (lead), Allen Xin Wang, Antonia Panescu, Lisa Xinyi Chen, Xinyu Wei, Ziyao Zeng;
Drug Discovery: Allen Xin Wang (lead), Antonia Panescu (lead), Jihang Chen, Sihan Jiang, Ziqing Wang, Siyi Gu;
Single-Cell Omics: Tianyu Liu (lead), Wenxin Long (lead);
Spatial Omics: Lisa Chen (lead), Tianyu Liu, Wenxin Long;
EHR: Xinyu Wei (lead), Siyu Chen, Xinyang Hu;
Genetics: Yueqian Jing (lead), Haoran Shao, Leqi Xu, Wangjie Zheng, Zhiyuan Cao;
Cross-domain:  Tianyu Liu (lead),  Allen Xin Wang, Xinyu Wei, Ada Fang;
Writing of the original draft: everyone;
Editing of the original draft: everyone;
Supervision: Tianyu Liu, Hongyu Zhao.

\section{Acknowledgments}

Tianyu Liu acknowledges the help from Dr. Minsheng Hao for helping us improve the quality of tasks. Tianyu Liu acknowledges the support from the Wu Tsai Institute at Yale for computing resources. (Also add fundings etc. to here.) This study was partially supported by the National Library of Medicine (NLM) of the National Institutes of
Health (NIH) under Award Number R01LM014604 (acknowledged by Prof. Hua Xu and Prof. Qingyu Chen).

\section{Conflict of Interests}

Need confirmation with all authors.

\bibliographystyle{unsrt}
\bibliography{sn-bibliography}

\newpage

\appendix
\counterwithin{figure}{section}
\renewcommand{\figurename}{Supplementary Fig.}
\renewcommand\thefigure{\arabic{figure}}  

\renewcommand{\tablename}{Supplementary Tab.}
\renewcommand\thetable{\arabic{table}}  

\section{Categories of AI Agents}
\label{append:catego}

\begin{table}[h]
\centering
\small
\resizebox{\textwidth}{!}{\begin{tabular}{lccccc l}
\toprule
\textbf{Agent Name} & \textbf{Tool Use} & \textbf{Human-in-the-loop} & \textbf{Auto Research} & \textbf{Self-evolving} & \textbf{Domain} \\
\midrule
GPT 5.2 \cite{openaigpt522025} & N & N & N & N & General \\
Gemini 3 Pro \cite{gemini3prosystem2025} & N & N & N & N & General \\
Claude Sonnet 4.6 \cite{anthropicclaudesonnet2025} & N & N & N & N & General \\
ToolUniverse \cite{gao2025democratizing} & Y & N & N & N & Biomedicine \\
Codex \cite{openai_codex_chatgpt_2026} & Y & N & N & N & General \\
ClaudeCode \cite{anthropic_claude_code_overview_2026} & Y & N & Y & N & General \\
CellForge \cite{tang2025cellforge} & Y & N & Y & N & Biomedicine \\
STELLA \cite{jin2025stella} & Y & Y & Y & Y & Biomedicine \\
AutoBA \cite{autoba2024} & Y & N & Y & N & Biomedicine \\
Biomni \cite{huang2025biomni} & Y & N & N & N & Biomedicine \\
TxAgent \cite{gao2025txagentaiagenttherapeutic} & Y & N & Y & N & Clinical-only \\
Medea \cite{sui2026medea} & Y & N & Y & N & Medicine (more clinical) \\
ChemCrow \cite{bran2024chemcrow} & Y & N & Y & N & Chemistry \\
CACTUS \cite{mcnaughton2024cactus} & Y & Y & Y & N & Chemistry \\
ChemToolAgent \cite{yu-etal-2025-tooling} & Y & N & Y & N & Chemistry \\
DrugAgent \cite{liu2025drugagentautomatingaiaideddrug} & Y & N & Y & Y & Chemistry \\
LIDDiA \cite{averly-etal-2025-liddia} & Y & N & Y & N & Chemistry \\
DELTA \cite{unlu2025auditableagentplatformautomated} & Y & Y & Y & N & Chemistry \\
MRagent \cite{mragentperform} & Y & Y & Y & N & Genetics-only \\
\bottomrule
\end{tabular}}
\caption{Comparison of AI agents across capabilities including tool usage, human involvement, autonomous research ability, self-evolution, and domain specialization.}
\label{tab:agent_comparison}
\end{table}

To ensure the rigor and timeliness of our conclusions while keeping experimental costs under control, we selected representative methods for evaluation based on the capabilities of the AI agent. We define the categories of selected AI Agents from five different aspects, including the support of tool calling (Tool Use), the support of interaction between humans and AI Agents (Human-in-the-loop), the ability to automatically explore the best solution (Auto Research), the ability to learn from the interaction with environment for updating prompts and plans (Self-evolving), and Domain. Most of the selected AI agents can call different types of tools, while self-evolving is the less common feature. Nevertheless, our selected AI Agents have already covered flagship agent design ideas as well as leading methods in their source domain.

We also report the agent cost, agent version, and base model version in Supplementary File X. We tried to utilize the most advance version, and found that most agents do not consume high resources (more than \$20 per run), except Biomni. We also evaluate different modes of one agent, such as STELLA with two different modes memory-enhanced mode (STELLA (mem)) and basic mode (STELLA (basic)). 

\section{Differences between our framework and other benchmark studies}
\label{append:benchinfo}

\begin{table}[h]
\centering
\small
\resizebox{\textwidth}{!}{\begin{tabular}{l l c c c c c c c}
\toprule
\textbf{Benchmark} & \textbf{Science Domains} & \textbf{Stepwise} & \textbf{Extensibility} & \textbf{Difficulty} & \textbf{Diverse Input} & \textbf{Suggestions} & \textbf{Validity} & \textbf{Agent types} \\
\midrule
AstaBench \cite{bragg2025astabench} & Mainly CS & \xmark & \cmark & \xmark & \cmark & \xmark & \xmark & 3 \\
BixBench \cite{mitchener2025bixbench} & Comp Bio & \xmark & \xmark & \xmark & \xmark & \xmark & \xmark & 1 \\
BioAgent Bench \cite{fa2026bioagent} & Bio & \cmark & \xmark & \xmark & \xmark & \xmark & \xmark & 1 \\
BAISBench \cite{luo2025benchmarking} & Bio & \xmark & \xmark & \xmark & \xmark & \xmark & \xmark & 4 \\
CompBio Bench \cite{nair2026agentic} & Comp Bio & \xmark & \xmark & \cmark & \xmark & \xmark & \xmark & 1 \\
ChemBench \cite{mirza2025chembench} & Chem & \xmark & \xmark & \xmark & LLM-only & \xmark & \xmark & 1 \\
ScienceAgentBench \cite{chenscienceagentbench} & Comp Sci, Neuro & \xmark & \xmark & \xmark & \xmark & \xmark & \xmark & 3 \\
MedAgentBench \cite{jiang2025medagentbench} & Med & \xmark & \xmark & \xmark & \xmark & \xmark & \xmark & 1 \\
TerminalBench (2.0) \cite{merrill2026terminal} & Mainly CS & \xmark & \cmark & \cmark & \xmark & \xmark & \xmark & 4 \\
HLE \cite{ota2025causal} & General Science & \xmark & \cmark & \cmark & LLM-only & \xmark & \xmark & 1 \\
SDE \cite{song2025evaluating} & General Science & \xmark & \cmark & \xmark & LLM-only & \cmark & \xmark & 1 \\
Ours & General Science & \cmark & \cmark & \cmark & \cmark & \cmark & \cmark & 11 \\
\bottomrule
\end{tabular}}
\caption{Comparison of AI agent benchmarks across scientific domains and evaluation dimensions. The full forms of the following abbreviations for `domains" are as follows: CS-Computer Science, Comp Bio-Computational Biology, Bio-Biology, Chem-Chemistry, Comp Chem-Computational Chemistry, Comp Sci-Computational Science, Neuro-Neuroscience, Med-Medicine.}
\label{tab:agent_benchmark_comparison_simple}
\end{table}

Here we discuss the difference between the benchmark we proposed and the past benchmark studies. We consider eight dimensions to make a comprehensive comparison, and these dimensions include: scientific domains that the benchmark focus on (Science Domain), whether the benchmark support step-wise evaluation (Stepwise), whether the benchmark can be extended  with both question sets and agents (Extensibility), whether the benchmark has labels for difficult questions (Difficulty), whether the benchmark is to accept different types of inputs (Diverse Input), whether the benchmark offers suggestions to resolve the identified errors (Suggestions), whether the benchmark considers the validity of tasks (Validity), and the type of agents tested in this benchmark (Agent types).

Across existing benchmarks, there is a clear fragmentation along both scope and evaluation rigor. Domain-specific benchmarks such as BixBench, BioAgentBench, BAISBench, CompBioBench, and MedAgentBench focus on particular scientific areas (e.g., computational biology or medicine), but typically lack breadth across disciplines. In contrast, more general benchmarks like AstaBench, TerminalBench, HLE, and SDE aim for broader coverage but often prioritize simplified or LLM-only settings, limiting their ability to capture realistic scientific workflows involving tools, data heterogeneity, and multi-step reasoning. Methodologically, most prior work provides only partial support for stepwise verification and extensibility, with limited mechanisms for systematically controlling task difficulty or incorporating diverse input modalities. Furthermore, evaluation dimensions such as validity, robustness, and actionable suggestions are often either absent or weakly defined, and many benchmarks constrain evaluation to a narrow class of (e.g. LLMs) and do not consider enough confounders to investigate (e.g., costs, openness, etc.).

In contrast, our benchmark introduces a unified and comprehensive evaluation framework that addresses these limitations. It simultaneously spans multiple scientific domains (biology, chemistry, drug discovery, medicine, as well as cross-domain tasks) while maintaining full support for stepwise verification, extensibility, difficulty annotation, diverse input formats, and suggestion generation. Crucially, it emphasizes realistic agent evaluation, moving beyond LLM-only settings to support heterogeneous, tool-augmented, and multi-agent systems. Our design also incorporates rigorous validity checks and richer evaluation signals, enabling more reliable assessment of both intermediate reasoning processes and final outcomes. By supporting a significantly larger and more diverse set of agent types, our benchmark provides a scalable and generalizable testbed for studying AI agents across the full spectrum of scientific discovery tasks. Finally, we also evaluate several confounders which can affect the performances of AI Agents, including executability, costs, tools, openness, and robustness. Together, these factors contribute to a comprehensive and advanced evaluation framework for AI Agents.

\section{Prompt list}
\label{append: prompt list}

Task specific prompts are included in our released dataset. First, we provided the agent with the file path and allowed agents to read the file, and then specified the task definition as well as requirements to formulate agents' inputs. Examples of agent outcomes are provided in each main figure.

\newpage

\section{Supplementary Figures}

\begin{figure}[H]
    \centering
    \includegraphics[width=1\linewidth]{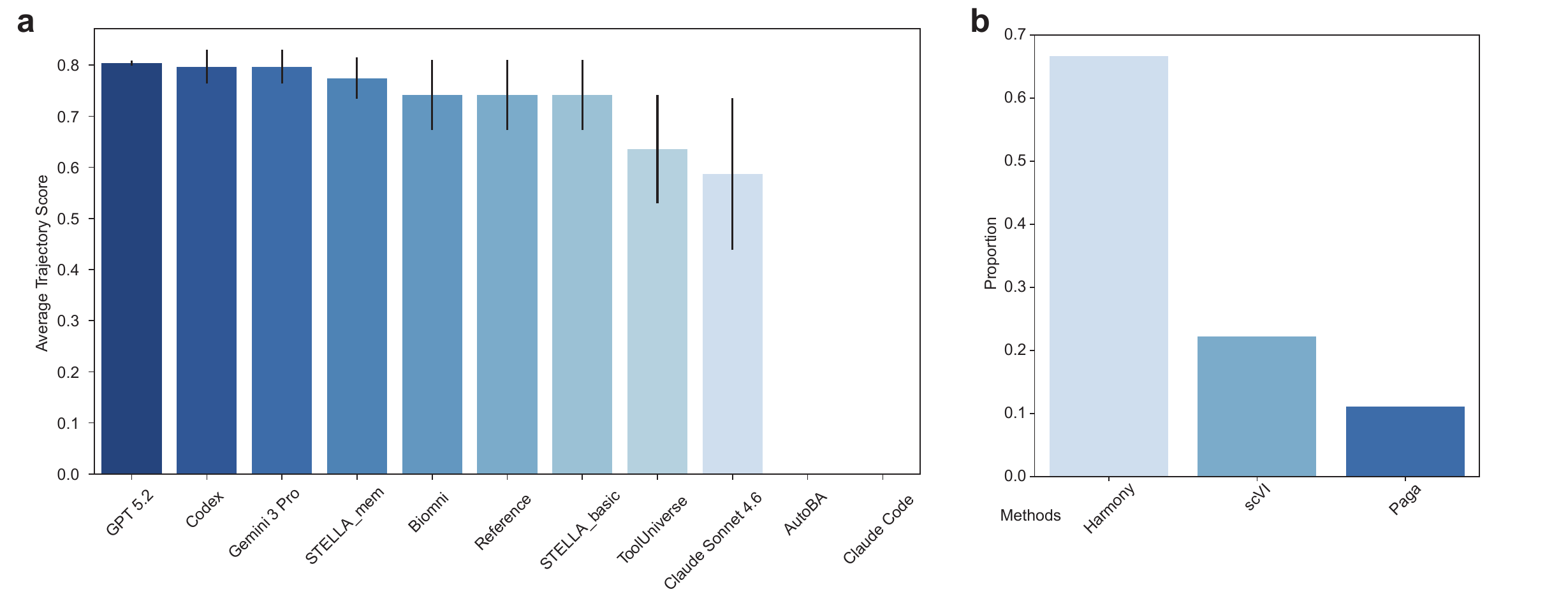}
    \caption{Performances of AI Agents based on the trajectory integration task. (a) Performances across agents. (b) Proportion of methods selected by agents in batch effect correction.}
    \label{supfig:traj_perf}
\end{figure}

\clearpage

\begin{figure}
    \centering
    \includegraphics[width=1\linewidth]{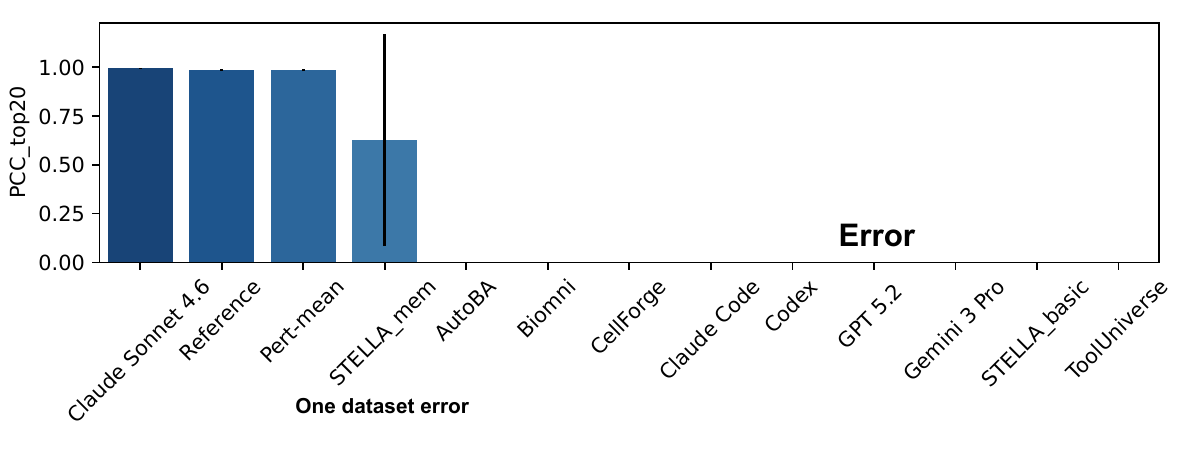}
    \caption{Perturbation prediction performances (PCC) evaluated based on the top 20 DEGs.}
    \label{supfig:pertdata_tp20}
\end{figure}

\clearpage

\begin{figure}
    \centering
    \includegraphics[width=1\linewidth]{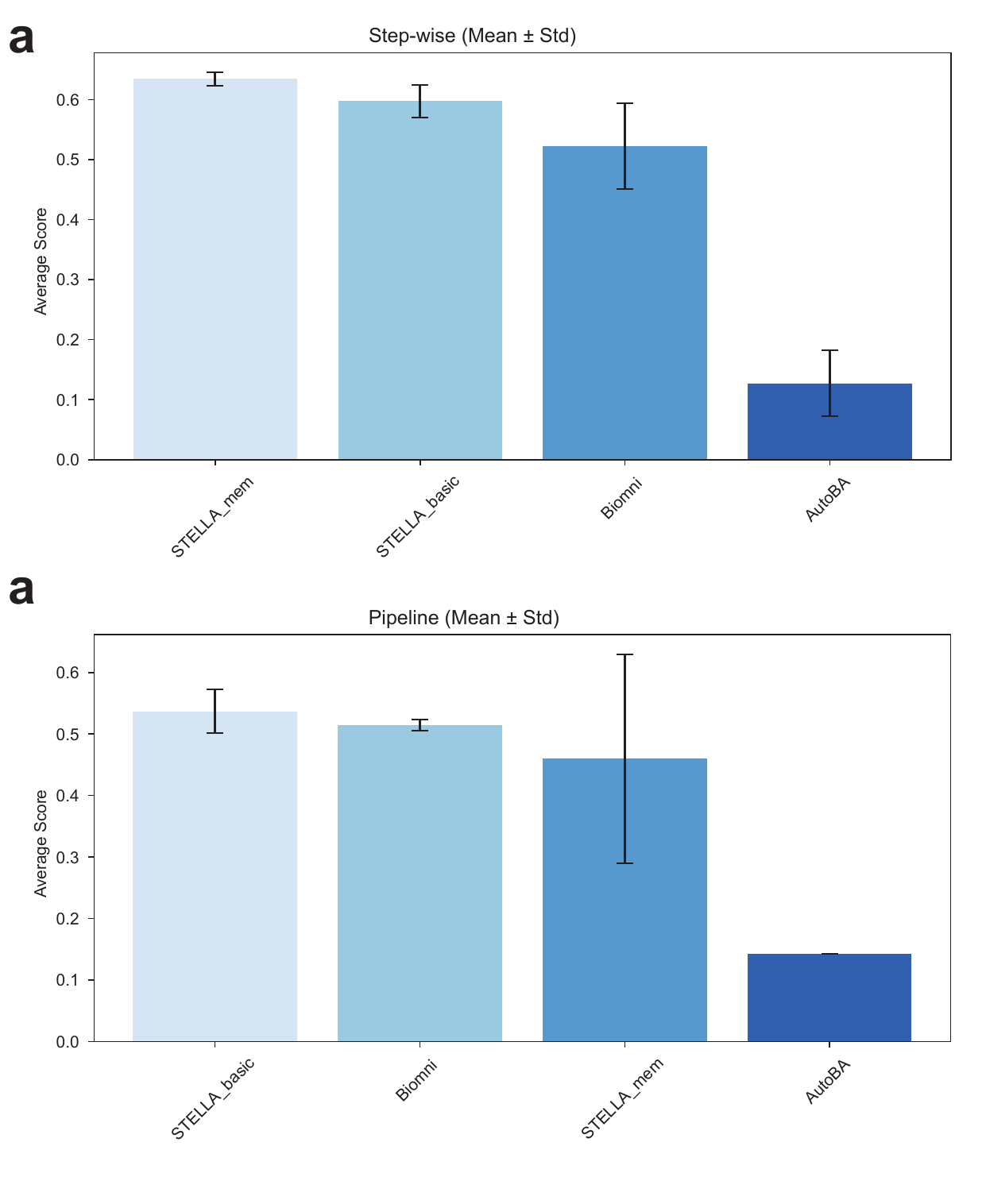}
    \caption{Stability Analysis for the workflow generation task based on three replicates. (a) Results of the step-wise mode. (b) Results of pipeline mode.}
    \label{supfig:stb}
\end{figure}

\clearpage

\begin{figure}
    \centering
    \includegraphics[width=1\linewidth]{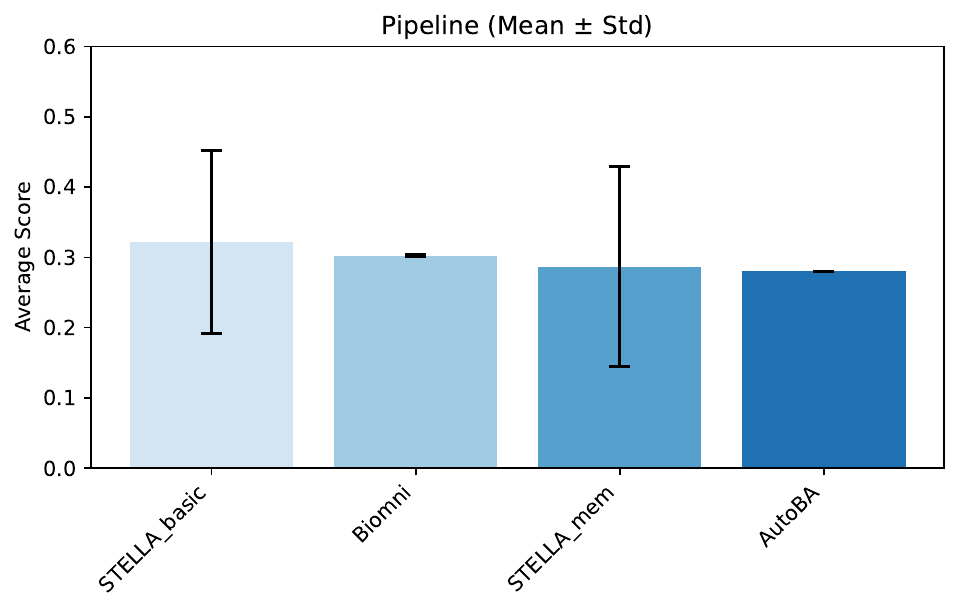}
    \caption{Stability Analysis for the workflow generation task based on three replicates. This figure shows the results of the pipeline mode based on Spatial Omics.}
    \label{supfig:sp_stab}
\end{figure}

\clearpage

\begin{figure}
    \centering
    \includegraphics[width=1\linewidth]{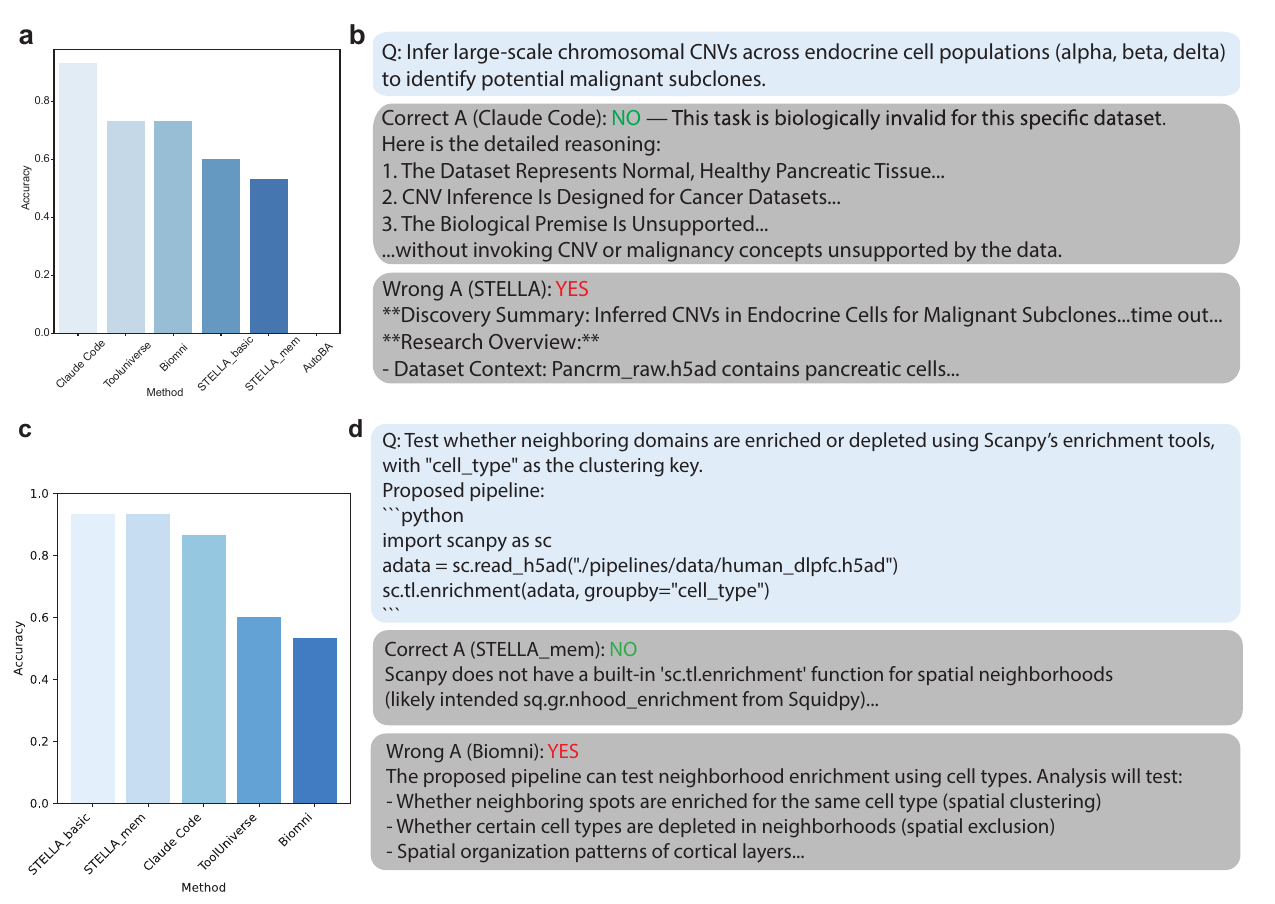}
    \caption{Results of testing the agents' abilities in identifying the valid tasks for omics data. (a) Performances of different agents for 15 tasks designed based on single-cell omics data. (b) One single-cell case study for the validity task, and we present one correct agent and one incorrect agent. (c) Performances of different agents for 15 tasks designed based on single-cell omics data. (d) One spatial case study for the validity task, and we present one correct agent and one incorrect agent.}
    \label{supfig:valid_task}
\end{figure}

\clearpage

\begin{figure}
    \centering
    \includegraphics[width=1\linewidth]{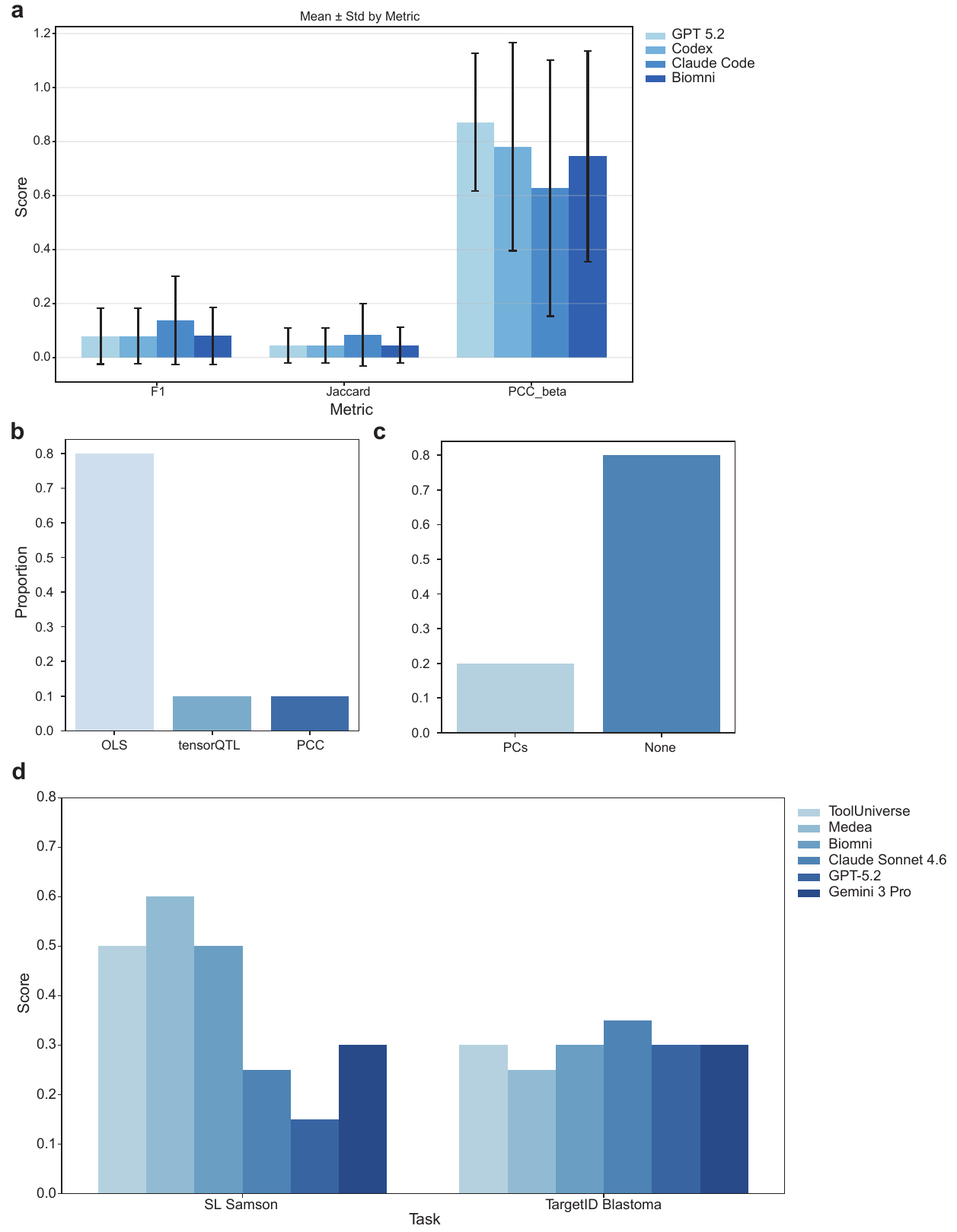}
    \caption{Results of cross-domain tasks. (a) AI agent performances for eQTL computation. (b) Proportion of methods selected by agents. (c) Proportion of covariates considered by agents. (d) Results of SL prediction and target identification.}
    \label{supfig:crossdomain}
\end{figure}

\begin{figure}[h]
  \centering
  \includegraphics[width=0.98\linewidth]{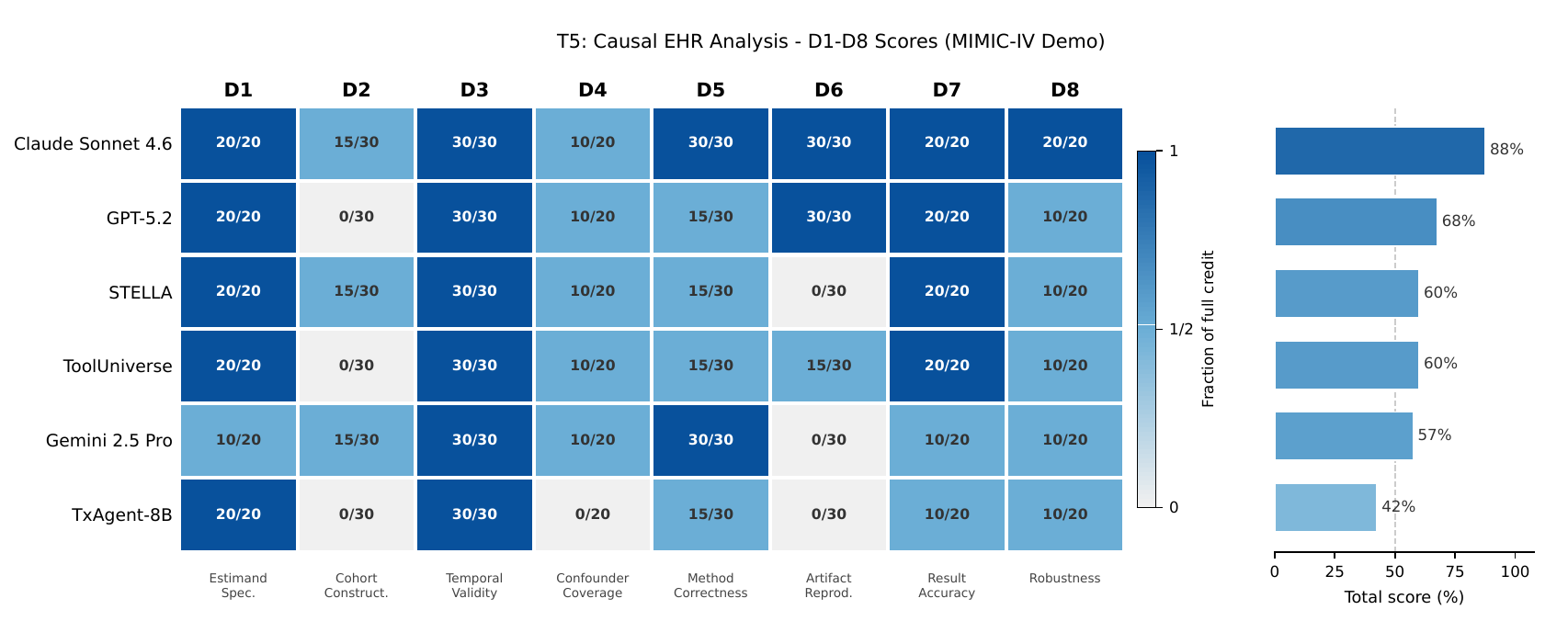}
  \caption{\textbf{T5 causal EHR analysis: per-agent, per-dimension scores (MIMIC-IV Demo).} Each cell shows the weighted score out of the dimension maximum (max\,=\,200 total); color intensity reflects the fraction of full credit. D3 (temporal validity) is perfect for all agents; D6 (artifact reproducibility) is the most discriminating dimension. Agents: Claude Sonnet 4.6, GPT-5.2, STELLA, ToolUniverse, Gemini 2.5 Pro, TxAgent-8B.}
  \label{supfig:t5_heatmap}
\end{figure}

\clearpage 

\begin{figure}
    \centering
    \includegraphics[width=1\linewidth]{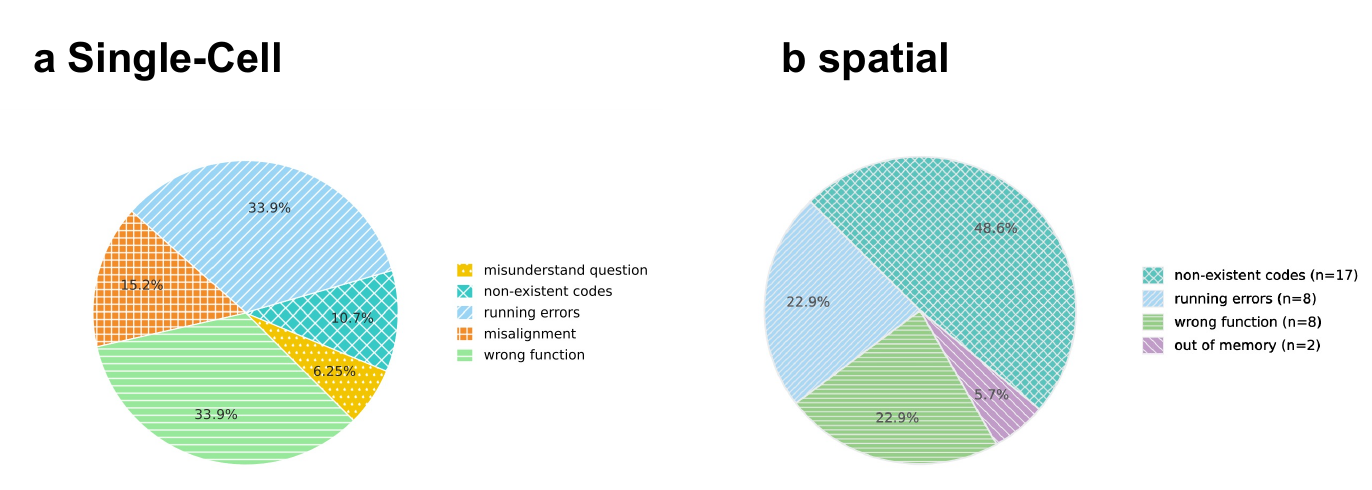}
    \caption{Error distribution for tasks based on omics data. (a) Error proportion for Single-Cell Omics. (b) Error proportion for Spatial Omics.}
    \label{supfig:error_pie}
\end{figure}

\end{document}